%% file: MAIN.tex
\documentclass[a4paper,fleqn]{cas-dc}

\usepackage[numbers,sort&compress]{natbib}

\usepackage{hyperref}
\PassOptionsToPackage{hyphens}{url}\usepackage{hyperref}

\usepackage{xurl}

\usepackage{empheq} 
\usepackage{xcolor}
\usepackage{ulem}
\usepackage{siunitx}
\usepackage{extdash}
\usepackage{comment}

\newcommand{\BS}{\color{magenta}}

\usepackage{amsthm}
\theoremstyle{definition}
\newtheorem{remark}{Remark}

\def\tsc#1{\csdef{#1}{\textsc{\lowercase{#1}}\xspace}}
\tsc{WGM}
\tsc{QE}
\tsc{EP}
\tsc{PMS}
\tsc{BEC}
\tsc{DE}

\begin{document}
\let\WriteBookmarks\relax
\def\floatpagepagefraction{1}
\def\textpagefraction{.001}
\shorttitle{Data-driven control of room temperature  and bidirectional EV charging using deep reinforcement learning}
\shortauthors{Svetozarevic et~al.}

\title [mode = title]{Data-driven control of room temperature and bidirectional EV charging using deep reinforcement learning: simulations and experiments}



\author[1]{Svetozarevic B.}[type=editor,
                        auid=000,bioid=1,
                      orcid=0000-0001-8921-3915]
\cormark[1]

\credit{Conceptualization, Methodology, Visualization, Writing - original draft, Supervision}

\address[1]{Urban Energy Systems Laboratory, Empa, Switzerland}

\author[1,2]{Baumann C.}
\credit{Methodology, Software, Data curation, Visualization, Writing - review \& editing} 

\author[2]{Muntwiler S.}[%
   ]

\credit{Methodology, Writing - review \& editing, Supervision} 

\address[2]{Institute  for  Dynamic  Systems  and  Control, ETH Zurich, Switzerland}

\author[1]{Di Natale L.}
\credit{Meth\-od\-ol\-o\-gy, Visulization, Writing - review \& editing} 

\author[2]{Zeilinger M.N.}
\credit{Review \& editing, Supervision} 

\author[1]{Heer P.}
\credit{Review \& editing, Project administration, Funding acquisition}


\cortext[cor1]{Corresponding author: \texttt{bratislav.svetozarevic@empa.ch} (B. Svetozarevic)}



\begin{abstract}
The control of modern buildings is 
a complex multi-loop  problem due to integrating renewable energy generation, transformation, and storage devices and connecting electric vehicles (EVs). On the other hand, it is a complex multi-criteria problem due to the need 
to optimize
overall energy use
while satisfying users' 
comfort.
Both conventional rule-based (RB)  and advanced model-based controllers, such as model predictive control (MPC),  
cannot fulfil the requirements of the  building automation industry  to 
solve this problem optimally
at low development and commissioning 
costs. 
RB controllers are difficult to apply in multi-loop settings, and MPC requires a building model, which is difficult to obtain for different buildings.

This work presents a fully data-driven, 
black-box pipeline to obtain an optimal control policy for a multi-loop 
building control problem
based on historical building and weather data, thus without the need for complex physics-based modelling. We demonstrate the method for joint control of room temperature and bidirectional EV charging 
to maximize the occupant thermal comfort and energy savings while leaving enough energy in the EV battery for the next trip. 
We modelled the room temperature with a  recurrent neural network
and EV charging with a piece-wise linear function. Using these models as a simulation environment, we applied a deep reinforcement learning (DRL) algorithm to obtain 
an optimal control policy. 
The learnt policy achieves on average 17\% energy savings over the heating season and 19\% better comfort satisfaction 
than a standard RB room temperature 
controller. 
When a bidirectional EV is additionally connected and a two-tariff electricity pricing is applied, the MIMO DRL policy successfully leverages the battery and decreases the overall cost of electricity compared to two standard RB controllers, one controlling the room temperature and another controlling the bidirectional EV (dis-)charging.
Finally, we demonstrate a 
successful transfer of the learnt DRL 
policy from simulation onto a real building, the DFAB HOUSE at Empa Duebendorf in Switzerland, achieving up to 30\% energy savings while maintaining similar comfort levels compared to a conventional RB room temperature controller over three weeks during the heating season.


\end{abstract}

\begin{graphicalabstract}
  \includegraphics[page=1,width=0.8\textwidth]{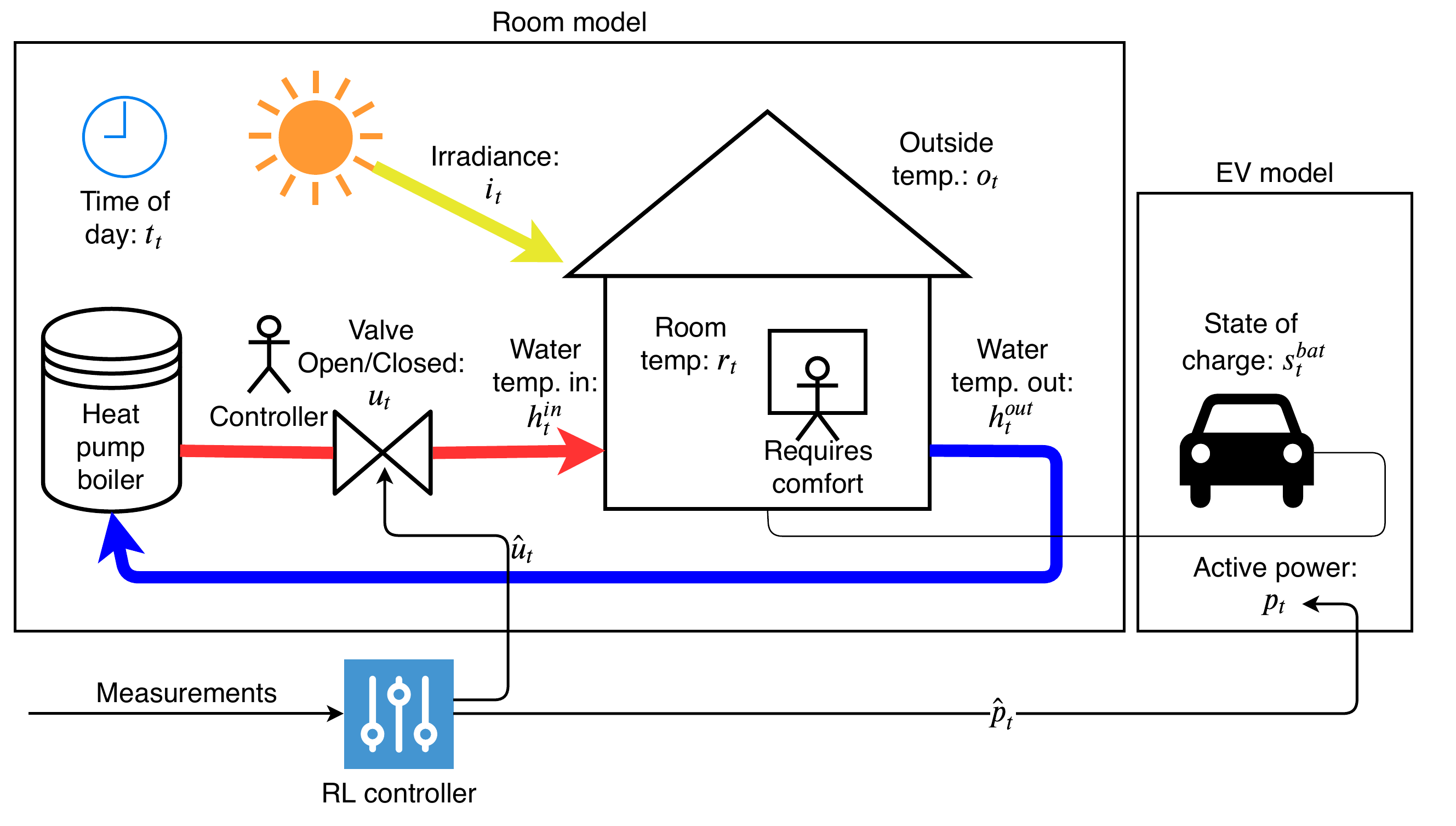}
\end{graphicalabstract}

\begin{highlights}
\item A black-box pipeline to generate optimal building control policies. 


\item The learnt policy 
simultaneously controls room temperature and EV charging.

\item Up to 30\% energy savings and better comfort compared to rule-based controllers.


\item Successful transfer of the  control policy 
from simulation onto the real building. 

\item The approach is suitable for other multi-loop building control problems. 
\end{highlights}


\maketitle

\input{Sections/01_Introduction.tex}

\input{Sections/02_Case_study}

\input{Sections/03_Methodology.tex}

\input{Sections/03b_Reinforcement_learning}

\input{Sections/05a_Simulation_results}

\input{Sections/05b_Experimental_results}

\input{Sections/06_Conslusions_and_discussion}

\printcredits

\input{Sections/10_Acknowledgement}

\section*{Appendix}

\appendix

\input{Sections/20_Appendix.tex}

\bibliographystyle{els-cas-templates/model1-num-names}

\bibliography{refs.bib}

\end{document}

%% file: Sections/01_Introduction.tex
\section{Introduction}

Buildings account for one-third of global primary energy consumption and one-quarter of 
greenhouse gas (GHG) emissions.  
Consequently, they have been identified as a critical 
element to enable climate change mitigation \cite{change2014mitigation}. When look\-ing at the energy use during a building's life-cycle, about 80\% of it stems from building operation \cite{ramesh2010life}.
However, over the last two decades, buildings have become much more complex 
to  operate optimally due to the integration of renewable energy generation, transformation, and storage devices \cite{shaikh2014review}. 
Also, due to the electrification of the mobility sector, electric vehicle (EV) chargers are 
installed 
in buildings, thus further increasing the control complexity \cite{huang2020coordinated}. Today, multiple energy-flows are possible within a single building, which gives rise to the need for system-wide optimal energy management. At the same time, the users' needs for comfort, such as indoor thermal and visual comfort, and having enough energy in the EV battery for the next trip, shall be satisfied. 

This multi-loop, multi-criteria control problem has raised specific needs within the building automation (BA) industry 
related to delivering optimal 
performance at low development and commissioning costs. 
In the following text, we first provide an overview of the BA industry requirements for an optimal controller for modern buildings. 
Then, we list the limitations of the current widespread rule-based (RB)
controllers and the advanced, state-of-the-art model-based controllers, of which Model Predictive Control (MPC) is the most famous representative.
We describe why both control methods fail to satisfy the current BA industry requirements. 
Following that, we motivate the potential of deep reinforcement learning (DRL) algorithms for the BA industry. We briefly review the current 
work on DRL applied to room temperature and EV charging control, and we close the Introduction with an overview of this work and a summary of main contributions.

\subsection{Current BA industry requirements}
\label{sec:BA_ind_reqs}

\textit{BA requirement I - 
Multi-loop control policy:}
Compared to the situation before the 2000s, renewable energy generation, transformation, and storage devices have now been vastly integrated into new or retrofitted buildings allowing for more energy-efficient and cleaner operations, in terms of GHG and, in particular, CO$_2$ emissions   \cite{chel2018renewable, chwieduk2003towards}. A typical set of these devices could include photovoltaic (PV) panels, battery storage, a heat pump, and thermal 
storage. 
Hence, the number of possible energy flows and the number of decision variables have increased. 
For example, electricity could be obtained either from the grid, a stationary battery, or PV panels. 
Similarly, when and which electricity source to use to charge an EV
depends on several factors, such as 
weather prediction and price of the electricity. 
Therefore, control of a modern building is a multi-input-multi-output (MIMO), i.e., multi-loop, energy management problem, which requires finding a control policy for several \textit{controlled  variables} simultaneously while considering several \textit{external factors} that influence it. 

\textit{BA requirement II - Building-EV coupling:}
The building-mobility sector coupling allows for more efficient control solutions than when these two sectors are addressed separately \cite{zhou2019energy}. For example, when the electricity price is low, the building management system (BMS) could decide to heat the room, charge the EV, or store it in a stationary  battery for later use. 
On the other side, this coupling also brings challenges. 
The charging of EVs causes additional energy consumption for a building, increasing its total -- and possibly peak -- energy consumption. 
Furthermore, most EV chargers start charging with full power as soon as an EV is connected. Therefore, if multiple EVs are charged at the same time in a neighbourhood, the aggregated demand can be very high, potentially causing energy dispatching and grid stability issues. 

A particularly interesting symbiosis between a building and an EV arises when the latter is bidirectional, i.e., the EV battery can be charged and discharged. In that case, the stored energy could be used as a source of electricity for a building 
\cite{liu2013opportunities}. In this case, the EV battery expands the capacity of the stationary battery, if one is installed. The difference with the stationary battery lies in its availability – the battery of a bidirectional EV is only available when the EV is connected to the building. 
Therefore, a BMS can use the battery of a bidirectional EV for energy management when the EV is connected. However, a BMS shall also ensure to charge the EV battery to a satisfactory level before the next trip.

\textit{BA requirement III - Occupants comfort:} In developed countries, people spend on average 80-90\% of their time indoors. Therefore, the influence of building systems on occupants' well-being is deemed 
critical \cite{park2018comprehensive}. 
Consequently, occupants put more and more stringent requirements for comfort to facility managers, which is passed indirectly to the BA industry. Therefore, the value of a building controller is not only measured in terms of saved energy but also how comfortable the indoor  environment is to the occupants. 

\textit{BA requirement IV - Transferability:} Buildings differ \linebreak 
from each other in terms of \textit{construction properties} (e.g. \linebreak
floor layout, geometry, materials used, age), \textit{installed \linebreak building services} (e.g. heating, ventilation, and air-condi\-tion\-ing (HVAC) systems), \textit{outside conditions} (climatic region, orientation), and \textit{occupancy profiles}. Therefore, an ideal building controller shall be able to provide optimal 
performance not only for the building it is designed for, but also for other similar buildings. 
If the engineering effort to apply such a controller to a similar building is small or negligible in terms of expert knowledge and time required, then the controller is considered transferable. 


\textit{BA requirement V - Adaptability and continuous commissioning:} The dynamics of a building can change significantly during its lifetime for several reasons, such as a retrofit, 
a change in the occupancy profile, 
or ageing. An ideal building control shall detect a change in the building operation performance, e.g. if a building starts to consume more energy than it used to, and readjust its parameters, i.e. adapt to the new situation. This capability of a controller is also called continuous commissioning \cite{salsbury2005survey, verhelst2017model}. 

Overall, the control of a  modern building is a complex 
MIMO control problem with the objective to provide the desired thermal comfort to the occupants and simultaneously ensure the EV is charged to a satisfactory level for the next trip, all while minimizing the overall energy consumption to reduce the costs. 


\subsection{Limitations of 
RB controllers}
\label{sec:limitations_RB}

Traditionally, more than 90\% of  industrial BA controllers are RB, such as bang-bang or pro\-port\-ional-integral-de\-riv\-a\-tive (PID) controllers. They have fixed predefined rules, simple architectures with straightforward implementation, \linebreak
and several parameters with clear guidance on how to tune them. Even though RB controllers (RBCs) are widely adopt\-ed in the BA industry, there are several limitations on their use for achieving optimal control of modern buildings. 


\textit{RBCs limitation I - Difficulty to achieve system-wide optimal performance:}  RBCs are suitable for single output control loops, whether single-input-single-output (SISO) or \linebreak
multi-input-single-output (MISO). 
Therefore, applying these single-output controllers to solve a multi-output (MIMO) control problem is a challenging and often infeasible task in practice as MIMO systems typically have dependencies between their sub-systems that cannot be neglected \cite{salsbury2005survey, stluka2018architectures, samad2020industry}. A MIMO system cannot be typically addressed as a collection of individual SISO/MISO systems \cite{skogestad2007multivariable}. For example, the temperature of the thermal storage determines the available heating capacity over the next couple of hours, while the output heating power of a heat pump connected to the thermal storage determines at what pace the temperature of the storage could be increased. A similar analogy could be drawn for the EV battery and its charging and discharging power. Indeed, optimal control of MIMO systems requires applying advanced MIMO control techniques \cite{skogestad2007multivariable}. 

\textit{RBCs limitation II - Absence of optimality guarantees:} Even for single-output problems, manual tuning cannot provide optimality guarantees: 
control experts could tune the RBC, in particular PID, to provide close-to-optimal regulation performance by looking at the overshoot, rise time, stability margins, and disturbance rejection, but there is no mathematical optimization involved in the tuning of the parameters. Therefore, most of the RB controlled loops in buildings perform sub-optimally \cite{salsbury2005survey, stluka2018architectures, samad2020industry}.

\textit{RBCs limitation III - Difficulty to include prediction rules:} RBCs do not typically involve any prediction rule. A prediction rule could be defined, for example, for pre-scheduled dynamic comfort bounds, which change between narrow, e.g. [\SI{22}{\celsius},   \SI{24}{\celsius}], and wider, e.g. [\SI{20}{\celsius},  \SI{26}{\celsius}] constraints. Such dynamic bounds are typical for office buildings, where wider comfort bounds are allowed outside of office hours 
to save energy. 
However, as an RBC would react to the change of the bounds only at the time of their change, 
such control will violate comfort. On the other hand, a predictive controller would pre-heat the room for some time before the narrow bounds need to be reached. 
Defining and tuning the prediction rule in an RBC would require experimenting with a building and determining the time dominant constant of a particular room 
so that the pre-heating interval could be defined precisely. 
However, this interval depends on the  day of the year, the state of the room, i.e.,  accumulated heat in walls, and weather prediction. Hence, determining it precisely for all combinations of these parameters over the year is a challenging task \cite{stluka2018architectures}.

\textit{RBCs limitation IV - Absence of self-adaptation:} 
RBCs need to be re-tuned after a change in building dynamics to regain the previous performance, which requires expert knowledge and incurs costs \cite{salsbury2005survey}. (See \textit{BA requirement V})

Overall, RBCs fail to satisfy all the needs of the BA industry for  an efficient way of obtaining an optimal controller for a modern building 
-- they can only provide sub-optimal performance 
and require expert knowledge during commissioning and maintenance.    

\subsection{Limitations of MPC controllers}
\label{sec:limitations_MPC}

Advanced controllers, on the other hand, in their classical and non-adaptive form, can overcome the first three limitations of RBCs. The most well-known representative of this type of controllers is 
MPC, which can calculate optimal MIMO control signals for several steps ahead while respecting the state, input, and/or output constraints. However, the performance of an MPC controller strongly depends on the quality of the underlying building model used to develop this controller. A building model of a poor quality, which does not represent well the true building dynamics, e.g. a simple grey-box model with some generic parameters, will lead to unacceptable control performance \cite{privara2013building}. On the other hand, obtaining a high-quality building model is a complex and time-consuming task requiring expert knowledge. Therefore, the costs of developing and implementing an MPC controller are justifiable only for well-defined systems, where the same controller could be used on many instances of the same system. However, as buildings differ substantially from each other, the costs of developing and deploying MPC controllers outweighs the cost benefits, and, therefore, classical versions of MPC controllers have not yet been widely adopted in the BA industry \cite{privara2013building, jain2018learning, serale2018model}.

Over time, stochastic \cite{oldewurtel2012use}, robust \cite{xu2010robust}, and adaptive \cite{tanaskovic2017robust} MPC controllers have been developed to address or circumvent the need for a high-quality building model. Even though some directions are promising, in particular those of adaptive MPC controllers with online system identification \cite{tanaskovic2017robust}, they have only been applied to single-zone temperature control problems and validated in simulation. Validation 
on real buildings and  solving of more complex building problems is needed for these methods to be accepted by the BA industry. 

\subsection{State-of-the-art data-driven RB and MPC controllers}

In recent years, due to the increased availability of stored sensor and actuator data in buildings, researchers have started exploiting the information contained in this past data to develop improved building controllers. 
Two trends can be observed: first, using data to improve classical control strategies, such as RB and MPC, and second, applying pure data-driven methods from the machine learning (ML) domain and adapting them to building control. 

The first direction, data-driven autotuning of RBCs, even though interesting from the industry perspective due to potential direct applicability, has not yet been widely addressed in the literature -- only some recent preliminary results exist \cite{khosravi2019machine, khosravi2021performance}.

Considerably more work has been published in the domain of learning-based MPC (LB-MPC) recently \cite{aswani2013provably, aswani2011reducing}. The most widely spread approach is to model the building dynamics with a neural network (NN) and use it as a model in the MPC framework. 
However, as NNs are non-linear models, the main challenge is to use them in a linear or convex fashion so that efficient solvers could be applied. One option is to  design a NN that can be used within MPC by constraining the output of the model to be convex with respect to the control inputs \cite{2018arXiv180511835C}. Besides NNs, Jain et al. \cite{jain2018learning} uses Gaussian processes to learn a model, which is then used within MPC. Another approach uses random forests for modelling 
\cite{smarra2018data}. Recently, it has also been validated experimentally, and preliminary results are promising \cite{bunning2020experimental}.
Even though initial results on data-driven MPC are promising, what is missing is the discussion on the scalability and transferability of these approaches across different buildings (see \textit{BA requirement IV}).

\subsection{Potential of DRL for building control}
\label{sec:DRL_potentials}

In terms of pure data-driven ML methods, reinforcement learning (RL), and in particular DRL, 
have emerged in recent years as approaches that can fulfil all the requirements 
for 
modern building control. 
Even though RL was established in the 1960s \cite{sutton2018reinforcement},  complex problems remained out of reach until recently, when people started using NNs within the RL framework. 
Together with the increased availability of large data sets and extensive computational power on demand, this led to the popularisation of DRL methods and the demonstration of successful solutions for complex real-world problems \cite{lecun2015deep, arulkumaran2017deep}. 
Mnih et al.  showed that DRL algorithms could achieve human-level or even super-human level intelligence in playing Atari games \cite{mnih2015human} and the game of Go \cite{silver2017mastering}. Since then, other problems have been solved at human or super-human levels in image recognition \cite{lecun2015deep}, natural language processing \cite{young2018recent}, and  medicine \cite{esteva2019guide}. Motivated by these  achievements in DL and DRL, building and control engineers started applying these methods to 
building control \cite{wang2020reinforcement, vazquez2019reinforcement,mason2019review}. There are several reasons why DRL is a promising framework to fulfil all the \textit{BA} industry \textit{requirements} 
for control of modern buildings. 

\textit{DRL potential I:} 
DRL algorithms operating on a continuous state space, such as deep deterministic policy gradient (DDPG) \cite{lillicrap2015continuous}, can  learn a continuous control policy to maximize a given reward function through interactions with the environment. As DDPG requires many interactions with the building, this is not feasible in practice, and people have to rely on models to learn the optimal policy. There are no particular requirements on the underlying model, such as convexity condition, as needed in MPC. 
As a building model, one could use any kernel-type model. NNs are particularly popular as they can capture the non-linear dynamics of the building \cite{ruano2006prediction,mustafaraj2011prediction}. After fitting the model to the past data, it is 
used as the simulation environment in the RL framework.  


\textit{DRL potential II:} There are no restrictions on how the reward function could be defined. Not only a single criterion but also multi-criteria reward functions and trading off requirements could be used. Hence, complex MIMO control policies could be obtained at once (see \textit{BA requirements I, II, and III}). 

\textit{DRL potential III:} Once the method is working for a certain room or building, it could also be applied to other rooms or buildings. 
The main part of the algorithm could be reused directly, thus demonstrating the transferability of the method (see \textit{BA requirement IV}). This problem is known as transfer learning, and it has been already considerably addressed in 
general reinforcement learning \cite{taylor2009transfer}. However, only limited prior work was published recently on the transferability of DRL algorithms for building control \cite{xu2020one}. 


\textit{DRL potential IV:} Finally, if updated with the newly received measurement data, the DRL algorithm could be updated online to adapt to the new building dynamics, thus fulfilling the \textit{BA requirement V} \cite{mocanu2018line}. 

\begin{figure*}
  \centering
  \includegraphics[page=1,width=0.8\textwidth]{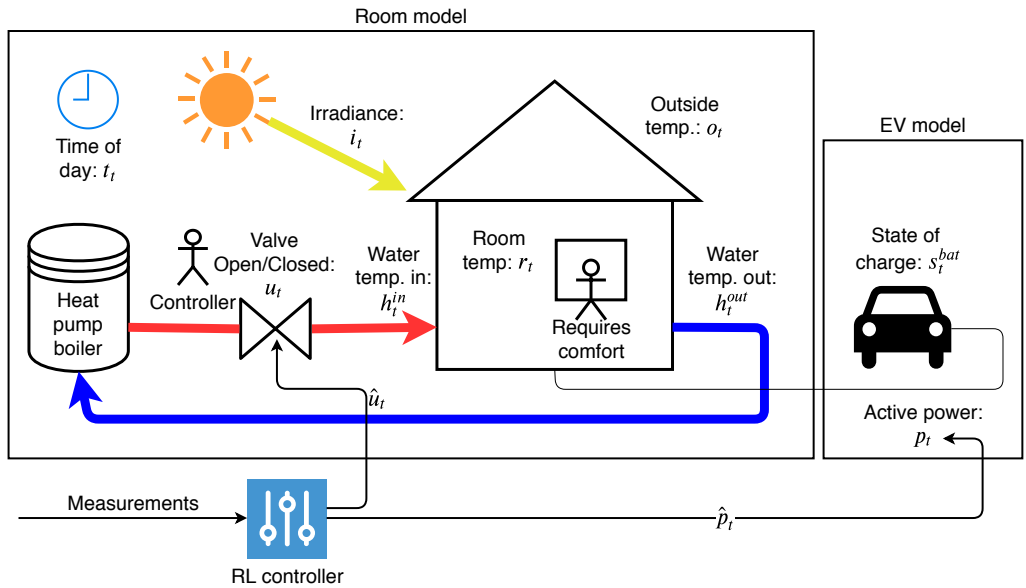}
  \caption{Overview of the room model, 
  bidirectional EV model, and 
  joint deep reinforcement learning controller}
  \label{fig:room_mod}
\end{figure*}

\subsection{State-of-the-art DRL-based room temperature and EV charging control}

Most previous works on RL and DRL 
consider either controlling the building
energy system, e.g. \cite{8060306, CHEN2018195, 2017arXiv171102278C, 
AFRAM201796, bunning2020experimental, pr5030046, 7401112}, 
or EV charging, e.g. 
\cite{6102330, 7178338, 6695263, 8727484, 8335743, 7056534}. 
There are a few works that control both the charging of an EV and a building energy system, 
e.g. \cite{en11082010, 6892986, 6547831, 6596523}. In \cite{en11082010} for example,
a building equipped with PV, 
an EV and an energy storage system is considered as a smart grid system, but no temperature control is addressed. The authors of \cite{6892986} 
minimize the costs of electricity through improved operations of an HVAC system, an EV, a washing machine and a dryer. 
In \cite{6547831, 6596523}, one-day ahead planning is used for building control, including an EV supporting bidirectional charging.

\subsection{Novelty and contribution of this work}


In this work, we describe a fully black-box, data-driven, DRL-based method for the joint control of a room temperature and bidirectional EV charging (see Fig. \ref{fig:room_mod}). The main contributions of this work are the following. 

First of all, the proposed data-driven pipeline requires only historical data to learn an optimal building control policy, and thus avoids the need for complex physics-based modelling required to develop advanced, model-based controllers (see Sec. \ref{sec:limitations_MPC}) or expensive fine-tuning of the conventional, rule-based controller (see Sec. \ref{sec:limitations_RB}).
 As a black-box simulation environment that does not require any physics-based prior knowledge to train the policy, we use a Recurrent Neural Network (RNNs) model of the room thermal dynamics and a linear model of the EV battery. We applied Deep Deterministic Policy Gradient, which is a DRL algorithm in the continuous domain, to learn the control policy. Hence, this pipeline is a cost-effective way to obtain an optimal MIMO building control policy by only using available historical data of a building. 

Secondly, we use the historical data from a real building, the DFAB HOUSE at Empa Duebendorf in Switzerland to  obtain a close-to-reality simulation environment. We analyse the simulation results of the DRL policy in a heating season in terms of energy savings and occupant comfort and showed that it delivers better performance than a standard industrial RB controller. Furthermore, we considered an extended problem when bidirectional EV is connected to the building and the electricity price has two tariffs. We analysed the simulation results of the simultaneous control of room temperature and bidirectional EV (dis-)charging in terms of costs savings while minimizing the comfort violations for the desired comfort bounds and providing enough energy to the EV battery for the next trip. 
The obtained DRL-based control policy showed better performance compared with two standard industrial RB controllers -- one for temperature regulation and another for EV (dis-)charging. 

Thirdly, we validated experimentally the learnt DRL policy for room temperature control during the heating season. The DRL policy was directly  transferred from simulation onto the real building, the DFAB HOUSE, and it was successfully regulating the temperature from the initial time of deployment, achieving  up to 30\% energy savings and better comfort satisfaction compared to a conventional, rule-based controller. 

Fourthly, we discuss throughout the paper the potential of this approach to satisfy all the \textit{BA Requirements (I-V)}.

\subsection{Structure}

The paper is structured as follows:
In Section \ref{sec:case_study}, the case study used to showcase the proposed data-driven building control methodology and the data collection process are described. 
In Section \ref{sec:Methodology}, we present the methods used to model the room temperature and the SoC of the bidirectional EV. Further, we describe the definition of the RL environment and the reward functions for  two different problems: i) room temperature control and ii) joint control of the room temperature and  bidirectional EV charging. 
The simulation and experimental results are illustrated in Section \ref{sec:results}. 
Finally, Section \ref{sec:conclusion_and_Discussion} provides an overview  and concluding remarks of this work, as well as directions for future research.

%% file: Sections/02_Case_study.tex
\section{Case study 
and data collection}
\label{sec:case_study}

\begin{figure}
  \centering
  \includegraphics[page=1,width=1\columnwidth]{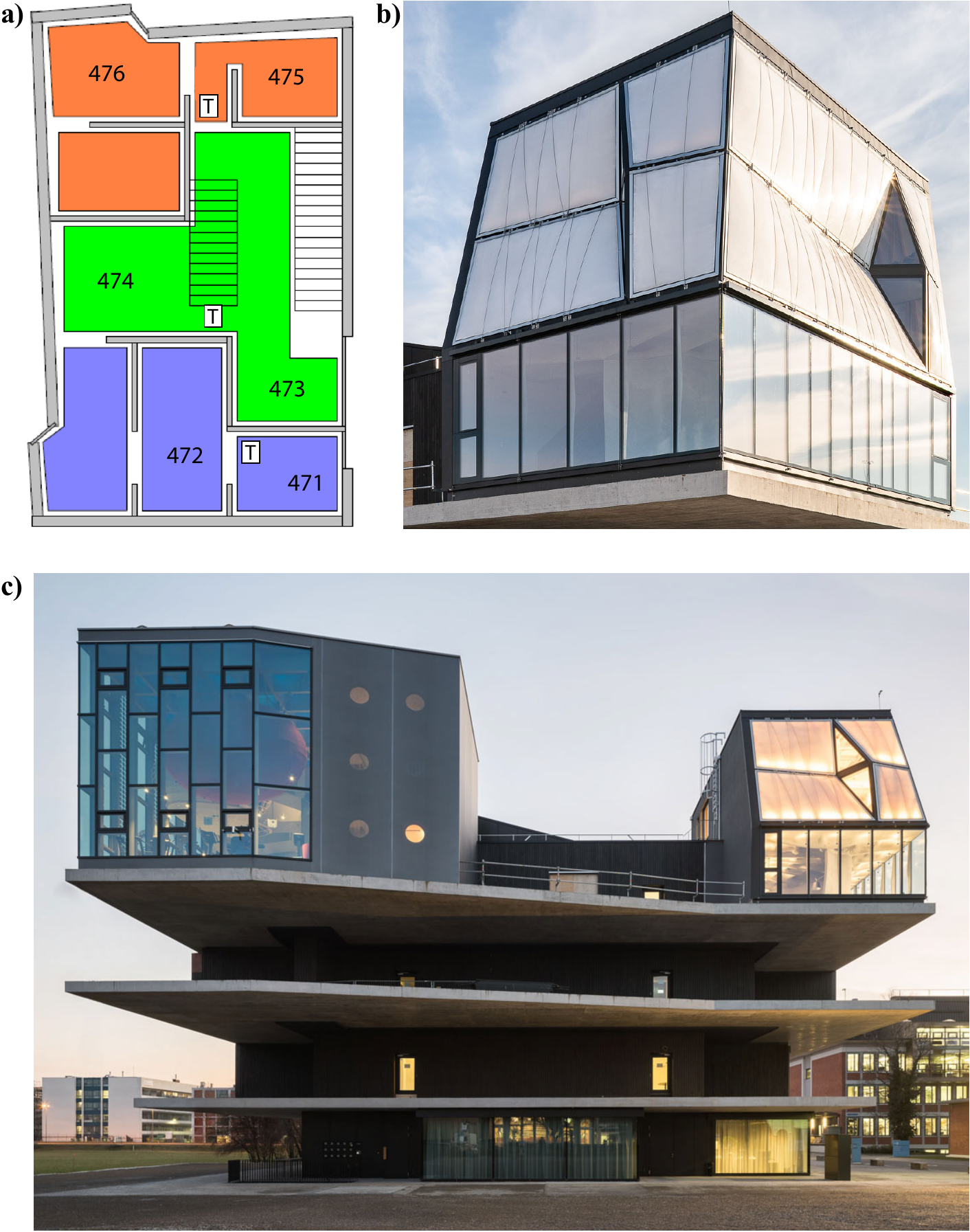}
  \caption{DFAB HOUSE.  a) The 2nd floor layout. The room temperature controller was developed for the room 471, on the bottom right. The locations of thermostats are marked with a square containing a "T". b) External view. c) NEST demonstration building at Empa, Duebendorf, Switzerland (Credits:  R. Keller \cite{dfab_img})}
  \label{fig:dfab_rooms}
\end{figure}

The DFAB HOUSE, a three-storey residential building of the Empa demonstrator NEST in Duebendorf in Switzerland \cite{noauthor_dfab_nodate}  (Fig. \ref{fig:dfab_rooms}). NEST  \cite{noauthor_nest_nodate} (Fig. \ref{fig:dfab_rooms}c) is a vertically integrated neighbourhood and a living lab. The DFAB HOUSE is  operational since March 2019 and the corresponding sensor and actuator data is collected at \SI{1}{\min} resolution. 
We chose one bedroom (room 471, Fig. \ref{fig:dfab_rooms}a) to apply our control algorithm. 
In this room, we can control the opening and closing of the valve that regulates the water flow into the floor heating system.
As a bidirectional EV was not available at the time of this work, we emulated it based on the past charging/\-discharging data of the stationary battery at NEST. 
For  information  on data preparation 
see Appendix \ref{sec:data_proc}.



%% file: Sections/03_Methodology.tex
\section{Methodology}
\label{sec:Methodology}

In the following subsections, first, the overview of the control problem and the corresponding model are provided. Then, the data-driven pipeline consisting of the data-driven modelling,  RL environment, and DRL algorithm is described.



\subsection{The control problem and model overview}


The overview of the system to be controlled is illustrated in Fig. \ref{fig:room_mod}. 
It consists of two parts: the room temperature model and the EV battery charging/discharging model. These two models are mainly independent, as they serve two different needs of the building occupants, namely to provide indoor comfort and enough battery capacity for the next trip, respectively. They are, however, linked through the overall building electricity demand. 
If the EV is being charged, the used energy indeed represents additional building energy demand. If the electric energy for heating/cooling is sourced from the bidirectional EV battery instead from the grid, then the overall building demand is reduced.

We can therefore formulate the 
control problem 
as: 
given the energy stored in the bidirectional EV battery, what would be the 
optimal room temperature control 
(heating or cooling) and 
optimal EV (dis-)charging strategy 
such that the overall costs for energy is minimized while satisfying the indoor comfort bounds and the minimum SoC of the EV at the moment of leaving. We assume that the EV leaves at 7:00 with a minimum of 60\% SoC and returns at 17:00 with  30\% SoC. The energy price is assumed to take a
standard two-stage tariff profile, with a high price between 8:00 and 20:00 and a low price outside of this interval.

\subsection{Data-driven modelling of the room temperature and bidirectional EV charging}
\label{subsec:modeling}

In this section, 
the model of the room temperature and the weather are described. Then, the two models are combined to obtain the final room temperature model, and we provide details on RNN architecture, model training, and hyperparameter tuning. Finally, we describe the bidirectional EV charging/discharging model.

\subsubsection{Room temperature model}
\label{sec:Room_model_definition}

We consider the temperature control of a single room (a single zone,i.e. room 471) at the DFAB HOUSE.
The room temperature $r_t \in \mathcal{T}^{room}$ depends on the outside temperature $o_t \in \mathcal{T}^{out}$, solar irradiance $i_t \in \mathcal{I}$,  
in-/out-flowing water temperature of the pipes $h_t^{in}, h_t^{out} \in \mathcal{T}^{heat}$,  and  the valve position $u_t \in \mathcal{U}$ (see Fig. \ref{fig:room_mod}). Here, the index $t$ denotes the time of the measurement. 
Since we will be using the room temperature model as a simulation environment for the RL agent, we need a model that predicts all uncontrollable (independent) variables.  
These are all of the above variables apart from the state 
of the valve $u_t$. Therefore, we define the state of the room $\mathcal{S}^{room}$ as the space of all non-controllable variables: $\mathcal{S}^{room} := \mathcal{S}^{full} \setminus \mathcal{U}
= \mathcal{T}^{out} \times \mathcal{I} \times \mathcal{T}^{room} 
\times (\mathcal{T}^{heat})^2 \times \mathcal{T}$ (see Table \ref{tab:spaces_ov} for definitions of all spaces used).

One way to solve the modelling task would be to fit the data with a multivariate time-series prediction model in an end-to-end fashion.
This would allow predicting the evolution of all the variables based on their past values.
Since the data collection at the DFAB HOUSE only started in March 2019, there was less than a year of operation and available historical data at the time of this work.
To make the most out of this limited amount of data, we took a few more considerations into account that led us to partition the room model into different sub-models. They are discussed in the subsequent sections.

\begin{remark}
The control framework described here could also be applied to different types of heating and cooling systems, where heating and cooling is provided by two different devices, e.g. an electric heater and an AC unit. 
\end{remark}

To get a smooth time variable, we use
$t^s_t = \sin(\tilde{t}_t) \in \mathcal{T}^s$ and 
$t^c_t = \cos(\tilde{t}_t) \in \mathcal{T}^c$, where 
$\tilde{t_t} \in \mathcal{T}^{lin}$ linearly goes
from $0$ to $2\pi$ during each day. To simplify the notation, we define $t_t \in \mathcal{T} := \mathcal{T}^s \times \mathcal{T}^c$ as the 
combined time variable. 
Note that one could also define the time in a linear fashion, numbering the time intervals during each day. However, this induces jumps at midnight from the last to the first interval. In other words, two extreme values are given to two adjacent intervals. Introducing the smooth sine and cosine time variables allows us to transfer the idea that these intervals are close to each other to the model.

\subsubsection{Weather model}

While there is a correlation between, e.g. the room temperature and outside temperature, the influence of the room temperature on the weather is non-existent. Therefore, to
avoid that output of the weather model depends on the room state variables, we train an individual
model of the weather. Such a model could be useful if no weather prediction data is available on site, but only past observed weather data could be taken as inputs. This model predicts the weather variables (outside temperature and irradiance)
based on the past values of those variables and the time of day. Let 
$\mathcal{W} := \mathcal{T}^{out} \times \mathcal{I}, 
w_t := (o_t, i_t)$ denote the combined weather data. The weather model is then defined 
as the \mbox{following} mapping:
\begin{equation}
    \label{eq:m_weather}
    \begin{split}
    m^{weather}: (\mathcal{W} \times \mathcal{T})^n &\rightarrow \mathcal{W}\\
    (w_{t-n+1:t}, t_{t-n+1:t}) &\mapsto \hat{w}_{t + 1}
    \end{split}
\end{equation}
Note that the weather model takes the $n$ previous values of the input series, i.e. $w_{t-n+1:t}$ and $t_{t-n+1:t}$, 
into account to produce the output. The "hat" notation denotes a prediction variable.

\begin{figure}
  \centering
  \includegraphics[width=1\columnwidth]{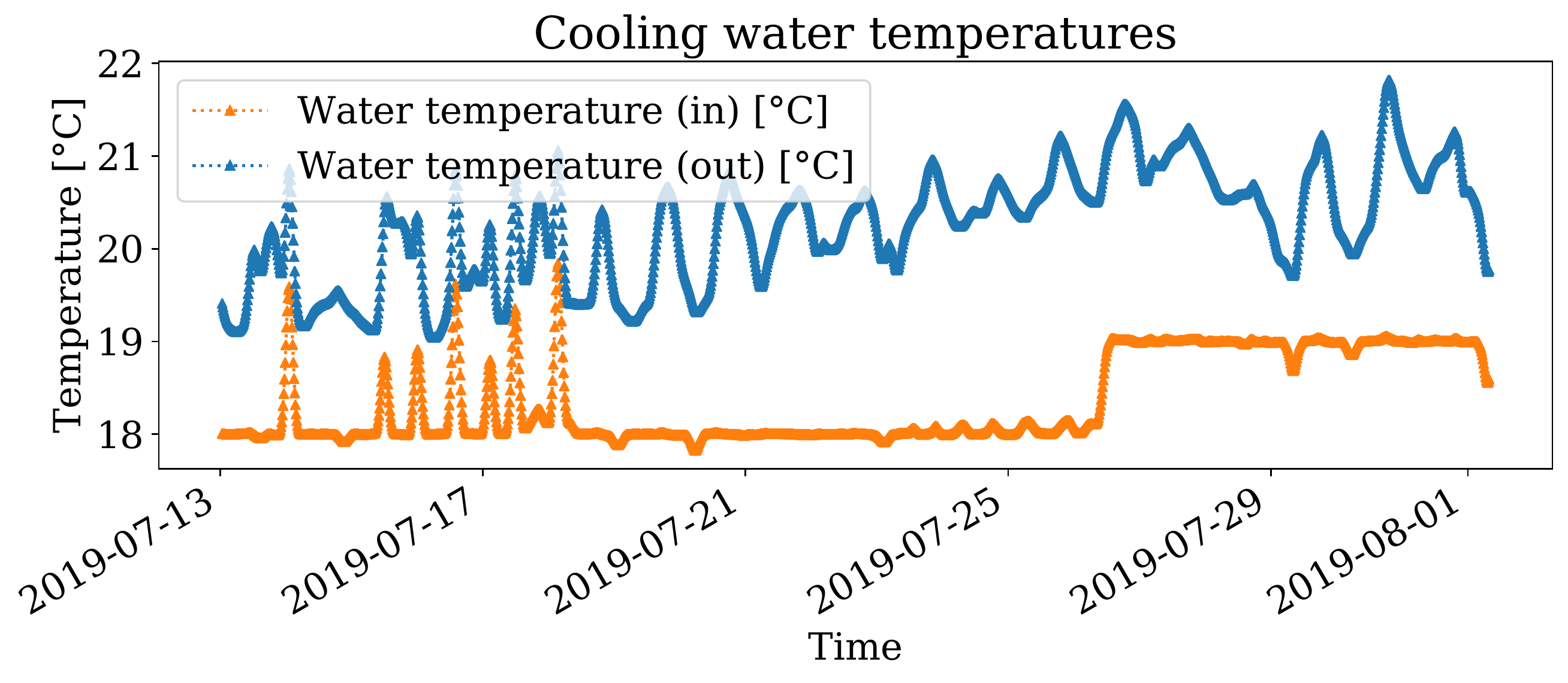}
  \caption{Inlet and outlet water temperatures of the cooling panel over two weeks in summer.}
  \label{fig:water_temp_1}
\end{figure}

The temperatures of the water entering and leaving the cooling/heating system over a few weeks in summer are shown in Figure \ref{fig:water_temp_1}. 
It can be seen that the water temperature coming from the heat pump is kept almost constant, but not always at the same level, which depends on the average outside temperature over a day. Since we are only interested in predictions with a rather short
the horizon of one day at most, we decided to use a constant predictor for the water temperature variables.  
While this is a valid assumption for the inflow temperature, the outflow temperature is much more dynamic. However, we retained this assumption for the sake of the simplicity of this model.

\begin{table*}
    \centering
    \begin{tabular}{|p{0.37\textwidth}|p{0.2\textwidth}|p{0.2\textwidth}|p{0.12\textwidth}|}
        \hline
        Variable name, Unit / Space name & Symbol & Space & Interval\\ 
        \hline\hline
        Outside temperature [\SI{}{\celsius}] &
        $o_t$ &
        $\mathcal{T}^{out}$        & [-15.0, 40.0] \\
        Room temperature [\SI{}{\celsius}] &
        $r_t$ &
        $\mathcal{T}^{room}$        & [10.0, 40.0]\\
        In- and Out-flowing
        heating water temp. 
        [\SI{}{\celsius}] &
        $h_t^{in}$, $h_t^{out}$ &
        $\mathcal{T}^{heat}$ &  
        [10.0, 100.0]\\
        Irradiance [\SI{}{\watt\per\meter\squared}] &
        $i_t$ &   
        $\mathcal{I}$ &            [0.0, 1300.0] \\
        Valve open / close state &
        $u_t$ &
        $\mathcal{U}$ &
        [0.0, 1.0] \\
        \hline
        Battery SoC [\SI{}{\percent}] & 
        $s_t^{bat}$ &
        $\mathcal{S}^{bat}$         & [0.0, 100.0]\\
        Active power applied to battery 
        [\SI{}{\kilo\watt}] &
        $p_t$ &
        $\mathcal{P}$ & [-100.0, 100.0]\\
        \hline
        Linear time of day [\SI{}{\hour}]& 
        $\tilde{t}_t$ &
        $\mathcal{T}^{lin}$ & [0.0, Inf]\\
        Sin and Cosine of time of day & 
        $t^s_t = \sin(\tilde{t}_t)$, 
        $t^c_t = \cos(\tilde{t}_t)$ &
        $\mathcal{T}^s$, 
        $\mathcal{T}^c$  & [-1.0, 1.0]          \\
        \hline
        Combined time of day &
        $t_t = (t^s_t, t^c_t)$      &
        $\mathcal{T} = \mathcal{T}^s \times \mathcal{T}^c$     &  \\
        Combined weather variables &
        $w_t = (o_t, i_t)$ & $\mathcal{W} = \mathcal{T}^{out} \times \mathcal{I}$  & \\
        Room state variable space &
        $s^{room}_t = (w_t, r_t, h_t^{in}$, $h_t^{out}, t_t)$ &
        $\mathcal{S}^{room} = \mathcal{W} \times \mathcal{T}^{room} 
        \times (\mathcal{T}^{heat})^2 \times \mathcal{T}$ & \\
        Full room variable space &
        $s^{full}_t = (s^{full}_t, u_{t+1})$  &
        $\mathcal{S}^{full} = \mathcal{S}^{room} \times \mathcal{U}$ &  \\
        Joint state space of room and battery &
        $s^{joint}_t = (s^{room}_t, s^{bat}_{t})$ &
        $\mathcal{S}^{joint} = \mathcal{S}^{room} \times \mathcal{S}^{bat}$ & \\
        \hline
    \end{tabular}
    \caption{Overview of variables used in the model and their corresponding mathematical spaces}
    \label{tab:spaces_ov}
\end{table*}

\subsubsection{Final room temperature model}
\label{subsec:final_room_T_model}

The final room temperature model $m^{rtemp}$ can now be defined. This model takes the previous $n$ values of the state variables in $\mathcal{S}^{full}$ and the controllable variable $u_t$ to predict the room temperature at the
next time step, i.e.:
\begin{equation}
    \label{eq:m_rtemp}
    \begin{split}
    m^{rtemp}: (\mathcal{S}^{full})^n &\rightarrow \mathcal{T}^{room}\\
    s_{t-n+1:t}^{full} = (s^{room}_{t-n+1:t}, u_{t-n+2:t+1}) &\mapsto \hat{r}_{t+1}
    \end{split}
\end{equation}
Note that we use $u_{t+1}$ to make the prediction. This is done deliberately since the model should give us the next state $\hat{r}_{t + 1}$ given the next control input $u_{t+1}$.

Putting everything together, we can now build the \textit{full} model of the room, $m^{full}$, by combining the previously defined sub-models: the weather model $m^{weather}$ \eqref{eq:m_weather},
the water temperature model $m^{wtemp}$ \eqref{eq:m_wtemp}, 
and the room temperature 
prediction model $m^{rtemp}$ \eqref{eq:m_rtemp},  as follows:
\begin{equation}
    \label{eq:m_full}
    \begin{split}
    m^{full}: (\mathcal{S}^{full})^n &\rightarrow \mathcal{S}^{room}\\
    s_{t-n+1:t}^{full} = (s^{room}_{t-n+1:t}, u_{t-n+2:t+1}) &\mapsto \hat{s}^{room}_{t+1}
    \end{split}
\end{equation}
with $s_{t}^{room} \in \mathcal{S}^{room}$ and 
$s_{t}^{full} = (s^{room}_{t}, u_{t+1}) \in \mathcal{S}^{full}$.
As mentioned previously, this model takes into account the $n$ previous values
of the input series $s^{room}_{t-n+1:t}$ and the same number  
of control inputs $u_{t-n+2:t+1}$ to compute the output.
By feeding each model the correct input
we can put together the desired output $\hat{s}^{room}_{t+1}$. 


\subsubsection{RNN model}
\label{ss:RNN_architecture}

\begin{figure}
  \centering
  \includegraphics[width=1\columnwidth]{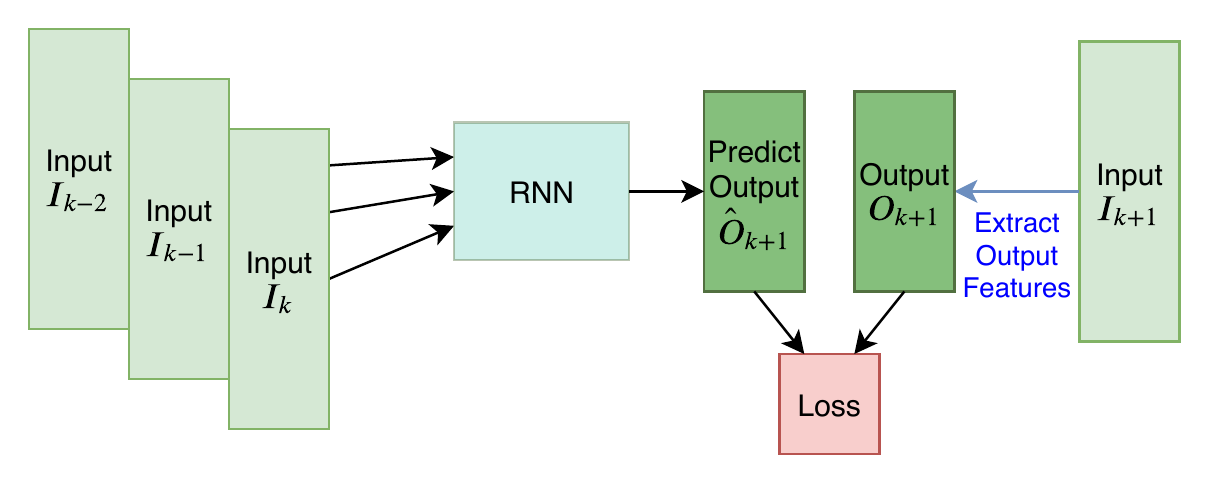}
 \caption{Simplified model structure for the case of     predictions based on three previous inputs ($n=3$). On the left, the model uses the provided inputs to make a prediction $\hat{O}_{k+1}$. On the right, it extracts the true output $O_{k+1}$ from the data, which can then be compared to the prediction to compute the loss and train the network. 
 }
  \label{fig:base_rnn_diag}
\end{figure}

RNNs are commonly used in time series predictions to capture its time dependencies and tendencies \cite{lipton2015critical}. Fig. \ref{fig:base_rnn_diag} illustrates how a single step prediction is made and this scheme is naturally expanded to multi-step predictions.
In that setting, part of the input is unknown and relies on the previous outputs of the model. It is then merged together with the known input part and fed to the RNN to build the next prediction. Repeating this procedure allows one to get predictions for any number of steps for weather and room temperature models. Note that in practice, we train the actual recurrent model to only predict the difference in the prediction state, not 
the absolute state. 

To optimize the loss, we use the ADAM \cite{adam} optimizer with a base learning rate $\eta$
to minimize the mean-square-error (MSE)
between the predictions and the ground truth.
The training of the model lasted for $n_{ep}$ episodes (see Table \ref{tab:hyp_rnn}). 
We also monitor the losses on the training and on the 
validation set to get an idea about the amount of overfitting.
The data used to fit the 
model is shuffled to avoid seasonal dependencies
between the data in consecutive batches.

\label{sss:hyp_tune}

The hyperparameters that are used to tune the recurrent models are listed in Table \ref{tab:hyp_rnn}.
There are a few more parameters that we choose heuristically, 
for example, a number of neurons in each recurrent layer. 
To compare the performance of the models trained with different hyperparameters, we use the following objective. We predict \SI{6}{\hour} (i.e. 24 timesteps of 15 minutes) into the future and take MSE between this prediction and the ground truth. For this process the validation data
is used. The main idea is to find a model that generalizes well
over multiple consecutive predictions 
and over unseen data.
For the actual optimization, a Tree Parzen Estimator \cite{hyper_tpe} is used, which is implemented in the Python library hyperopt \cite{bergstra2015hyperopt}.

\subsubsection{Bidirectional EV charging/discharging model}
\label{ssec:bat}

We use a stationary battery available at NEST in order to 
emulate the battery of a bidirectional EV.
This battery has a maximum capacity 
of \SI{96}{\kilo\watt\hour}
at a SoC of $100\%$. However, we will restrict
it to lie within the interval $[20.0\%, 80.0\%]$ for safety reasons (the details on the safety are discussed later in Section \ref{sec:ev_bat_env}).
Furthermore, we 
limit the charge and discharge rate to
\SI{+-100}{\kilo\watt}. Both stated maximum capacity 
and maximum (dis-)charging rate 
are also found in the following EV models: Tesla Models S and X \cite{dow_tesla_2020}, BMW i3 \cite{brodie_autoexpress_2018}, and Mercedes-Benz EQC \cite{taylor_better_2018}.

The change in SoC is modelled to be proportional 
to the active power applied, but the proportionality 
factor can be different for charging and discharging.
We also allow for a constant discharging rate when the battery 
is not used, i.e. if the applied active power is zero, the battery
 slowly decreases its SoC due to losses.
Let $s^{bat}_t \in \mathcal{S}^{bat} := [20.0\%, 80.0\%]$ 
be the SoC at time $t$, 
let $p_t \in \mathcal{P} := 
[-\SI{100}{\kilo\watt},$ 
$\SI{100}{\kilo\watt}]$ be the 
average active power from time $t-1$ to time $t$.
Finally let $\Delta s^{bat}_t := s^{bat}_t - s^{bat}_{t-1}$ be the
change in SoC at time $t$ compared to time $t-1$. 
Therfore, we model the change in SoC, or charging/discharging of the EV battery, as:
\begin{equation}
    \label{eq:pw_bat}
    \Delta s^{bat}_t(p_t) \approx \Delta \hat{s}^{bat}_t(p_t) :=\alpha_0 + \alpha_1 p_t + \alpha_2 \max \{0, p_t\}
\end{equation}
where $\alpha_i, i = 0,1,2$ are the variable coefficients
that can be fitted to the data using least squares. 
Finally, we
can define the battery model $m^{bat}$ as:
\begin{equation}
    \label{eq:bat_mod}
    \begin{split}
    m^{bat}: \mathcal{S}^{bat} \times \mathcal{P} &\rightarrow \mathcal{S}^{bat}\\
    (s^{bat}_t, p_{t+1}) &\mapsto \hat{s}^{bat}_{t + 1} := m^{bat}(s^{bat}_t, p_{t+1}) \\
            &:= s^{bat}_t + \Delta \hat{s}^{bat}_{t+1}(p_{t+1})
    \end{split}
\end{equation}

\noindent It models how the SoC evolves when an active power of $p_{t+1}$ is 
applied. We consider the model to be charging if the active power
is positive and discharging otherwise.

%% file: Sections/03b_Reinforcement_learning.tex
\subsection{RL environment}


In RL, an agent is learning a control policy through interaction with an environment. Let $\mathcal{S}$
be the state space and $\mathcal{A}$ be the action space and let $s_t$ and $a_t$ be the state and the action at time $t$, respectively. 
Then the environment denoted by $E$ is a 
mapping $E: \mathcal{S} \times \mathcal{A} 
\rightarrow \mathcal{S} \times \mathcal{R} \times \mathcal{B}^{term}$, where 
$(s_t, a_t) 
\mapsto (s_{t+1}, r_t, b^{term}_t)$, 
$r_t \in \mathcal{R}$ is the reward received at time $t$ and $b^{term}_t \in \mathcal{B}^{term}$ is the boolean value which indicates if the current episode is over. 
In this work, we trained our agents in an episodic framework, with a fixed episode length of
$l_{ep} := 48$. With one timestep corresponding to \SI{15}{\minute},
this corresponds to an episode length of \SI{12}{\hour}. The episode termination indicator $b^{term}_t$ is thus defined as true if $t = l_{ep}$, otherwise is false. 

In our case, we naturally use a transition model $m: \mathcal{S} \times \mathcal{A} 
\rightarrow \mathcal{S}$ with 
$(s_t, a_t) 
\mapsto s_{t+1}$,
which corresponds to the form of our room model. 
\noindent All we additionally need to build the RL environment is 
a reward function $r: \mathcal{S} \times \mathcal{A} (\times \mathcal{S}) 
\rightarrow \mathcal{R}$. 
The reward function returns the reward $r_t = r(s_t, a_t, s_{t+1})$
that agent gets when the selected action $a_t$ leads 
to a transition of the environment from
state $s_t$ to the next state $s_{t+1}$. The general objective of any RL agent is to maximize the reward. Therefore, if one wants to minimize a certain cost function, one possibility is to use the negative of the cost as reward. 

In the following sections, we  define the environment of our particular problem using the previously described room temperature and EV (dis-)charging models.

\subsubsection{Room temperature environment}

The model $m^{full}$ \eqref{eq:m_full} 
can predict all the variables needed to control the room temperature and is thus used as an environment in our case.
Therefore, we use $S^{room}$
as the state space for the RL environment and $A^{room} := U$, the space
of valve states, as the action space since that is what can be controlled directly. We also define $a^{room}_t := u_{t+1}$
as the action for the room temperature environment.

To initialize the environment in each episode, we sample an initial condition from the historical data in the 
database and we then use the model to simulate the behaviour of the
room under the agent's policy for the length of the episode. 
This episodic framework allows us to control the errors of the model, since
we know how well it performs.
Further, to incorporate stochasticity, 
a disturbance term $d(t)$ is added to the output of the deterministic model. 
This is assumed to help the agent find a
policy that is robust to disturbances in the model. 
Mathematically, we thus define the room temperature environment as:
Hence, the evolution of the states is defined as:
\begin{equation}
    \label{eq:room_evol}
    \begin{split}
    s^{room}_{t+1} &:= m^{room}(s^{room}_{t-n+1:t}, a^{room}_{t-n+1:t}) + d(t) \\
    \end{split}
\end{equation}
The disturbance itself
is modelled by an auto-regressive (AR) process that was fitted based
on the residuals of the NN model. This ensures that the disturbance 
is realistic, i.e. as seen in the past data.

The reward of the agent controlling the room temperature is defined as follows:
\begin{equation}
    \begin{split}
    r^{room}_t(s^{room}_t, a_t^{room}) &:= 
        -\overbrace{a^{room}_t \cdot |h_t^{in} - h_t^{out}|}^{\text{Energy usage}}
        - \alpha \cdot \overbrace{c^{pen}(r_t)}^{\text{Comfort  violation}}\\
        &= -e_t^{room} - \alpha \cdot c_t
    \end{split}
\end{equation}
\noindent where we defined 
$e_t^{room} := a^{room}_t \cdot |h_t^{in} - h_t^{out}|$ and 
$c^{pen}$ denotes the penalty function for room temperatures
that are outside the comfort bounds. 
The parameter $\alpha > 0$ determines
the weight of the temperature bound violation compared
to the energy usage.
The penalty function $c^{pen}$ is defined as follows:
\begin{equation}
    \label{eq:stop_c}
    c^{pen}(r_t) := 
  \begin{cases}
        0 & r_{min} \leq r_t \leq r_{max} \\
        r_{max} - r_t & r_t > r_{max} \\
        r_t - r_{min} & r_t < r_{min} \\
  \end{cases}
\end{equation}

\noindent Note that this function is always positive and increases
linearly with $r_t \rightarrow \pm \infty$ as soon as the temperature gets out of the defined comfort bound $[r_{min}, r_{max}]$.

\subsubsection{EV battery environment}
\label{sec:ev_bat_env}

To build the RL environment for the EV battery,
the battery model $m^{bat}$ (\ref{eq:bat_mod}) described in section \ref{ssec:bat} is used:
\begin{equation}
    \begin{split}
    E^{bat}: \mathcal{S}^{bat} \times \mathcal{A}^{bat} &\rightarrow 
    \mathcal{S}^{bat} \times \mathcal{R}^{bat} \\
    (s^{bat}_t, a^{bat}_t) &\mapsto (s^{bat}_{t+1}, r^{bat}_t)
    \end{split}
\end{equation}

The SoC of the battery at a given time $t$, $s^{bat}_t \in \mathcal{S}^{bat}$, is used as state of the environment and the space of the applied active power
$\mathcal{A}^{bat} := \mathcal{P}$ is used as action state, with the action defined as the active power 
$a^{bat}_t := p_{t+1} \in \mathcal{A}^{bat}$. Note that the subscripts do not match
since we defined $p_{t+1}$ as the active power applied from $t$ to $t+1$, but
this is also 
at time $t$.

Besides restricting the active power, 
we also want to restrict the SoC of the battery to lie within a 
certain range.
Since the battery model learnt from the data is piece-wise linear
and strictly increasing, it can be inverted and used to build a fallback controller. We implemented two functions in the fallback controller. First, the fallback controller prevents the SoC from falling
out of the previously defined safety range, $[20.0\%,80.0\%]$. The actions are not directly used but clipped using the safety guaranteeing
function $f^{safe}$ 
that will clip the chosen actions to the required range
for the constraints to be fulfilled.
More details on how this function is defined can be found in Appendix  \ref{appendix_constraining_bat_control}. As the constrained action needs to be fed to the learned model $m^{bat}$, the following is defined:

\begin{equation}
    \label{eq:bat_evol}
    \begin{split}
    s^{bat}_{t+1} &:= m^{bat}(s^{bat}_t, 
        f^{safe}(a^{bat}_t)) \\
    r^{bat}_t &:= -f^{safe}(a^{bat}_t)\\
    \end{split}
\end{equation}

Furthermore, the fallback controller achieves a specified SoC at the desired future time $t_{des}$ by restricting the battery to be charged at high 
power when the SoC is too low when approaching $t_{des}$. 
This makes it easy
to build an environment for RL: we can choose the reward as the negative active power applied per timestep and we do not need additional penalties contained within the reward that penalize SoCs outside
of the given bounds or not reaching the SoC goal at time $t_{des}$. 
In this way, we omit choosing a heuristic factor for balancing the energy used and the 
SoC constraint violation (see Appendix \ref{app:bat_env} for details on SoC constrain violation).

\begin{figure}
  \centering
  \includegraphics[page=1,width=1\columnwidth]{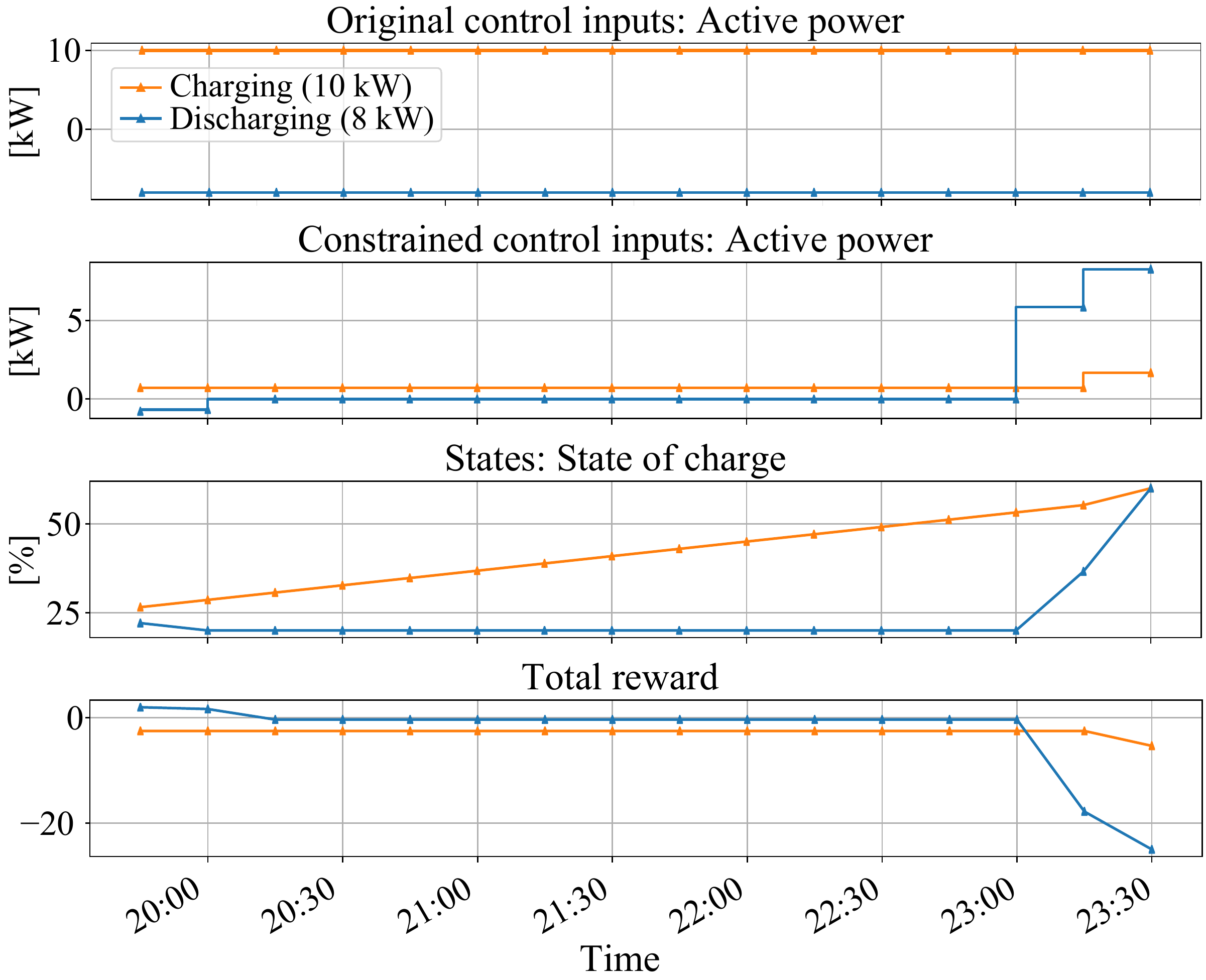}
  \caption{EV battery environment demonstration.} 
  \label{fig:bat_env}
\end{figure}

Figure \ref{fig:bat_env} shows how the resulting environment behaves under two different heuristic agents that
apply a constant action. One is discharging and the other 
is charging at a constant rate. Note that in this case, we chose $t_{des}$
as the end of the episode, i.e. $t_{des} = l_{ep} := 48$. 
One can see that the agent 
that constantly wants to discharge arrives at the minimum SoC after a few steps and needs to charge the battery at full capacity when approaching the
end of the episode. 
The safety controller built into the environment prevents the SoC from falling below the minimum and charges the battery 
before the end of the episode, even if the agents continue to discharge. 


\subsubsection{Joint room temperature and EV battery environment}

Since the joint environment consists of both the room  and
the battery environment, we combine them.
This means that the action space will be
$A^{joint} := A^{room} \times A^{bat}$ and the state space will similarly be combined as
$S^{joint} := S^{room} \times S^{bat}$, where $(s^{joint}_t, a^{joint}_t) 
\mapsto (s^{joint}_{t+1}, r^{joint}_t, b^{joint}_t)$. 

As both subsystems evolve independently, we simply use equations \eqref{eq:room_evol} and \eqref{eq:bat_evol} to compute their next state, that we can then concatenate to yield the next state of the joint system.
Since the reward was one-dimensional in both cases, we combined the two in a weighted sum as follows:

\begin{equation}
    \begin{split}
    \noindent r^{joint}_t(s^{joint}_t, a^{joint}_t) &:=\\
        -p(t) \cdot &\overbrace{(\alpha_{bat} \cdot e^{bat}_t + 
        e^{room}_t)}^{\text{energy usage}} -
        \alpha \cdot \overbrace{c^{pen}(r_t)}^{\text{comfort violation}}
    \end{split}
\end{equation}
\noindent 
where $p(t)$ denotes a suitable energy price function that may vary
over the course of a day, but is the same for different days.
Note that compared to the room temperature environment, in this case, we are no longer interested in energy minimization
but in price minimization. Maximizing thermal comfort remains also here.
Note, also, that  coefficient $\alpha$ here is introduced to balance out the consumption of the battery and the room, which have different scales.

\subsection{DRL algorithm}


In this work, we used
the Deep Deterministic Policy Gradient (DDPG) algorithm \cite{lillicrap2015continuous}.
It is model-free, off-policy, and uses an actor-critic setting. 
Unlike standard Q-learning, it naturally handles continuous state and action spaces, which was one of the main reasons this algorithm was chosen. This choice was also motivated by previous work using this algorithm, for example in
\cite{ddpg_app_1, 2017arXiv170708817V, 2016arXiv161200147X, ddpg_app_2}. 
There exists an implementation
of DDPG based on the python deep learning library Keras 
\cite{chollet2015keras} in another  library 
called Keras-RL \cite{plappert2016kerasrl}. 

Four neural networks are used within the DDPG algorithm: an actor (taking actions)
and a critic network (evaluating these actions) and corresponding target networks for each of them.
Note that the actor and its target network have the same architecture but different weights. 
The same applies to the critic and its target network.
In our case, a fully connected neural network with two layers of 100 units and the Rectified Linear Unit (ReLU) activation function was used for both the 
actor and the critic.  To perturb the actions
chosen by the actor network with exploration noise, 
an Ornstein-Uhlenbeck process (see e.g. \cite{Finch04ouprocess}) was used.
As for the RNN training in the modelling section, 
we used the ADAM  optimizer \cite{adam} to update the parameters
of the neural networks. 
The discount factor $\gamma$ was fixed to $0.99$.
Note that a few more hyperparameters, like the learning rate for the ADAM optimizer and
the number of training episodes, were adjusted manually. This could be avoided using automatic hyperparameter tuning,
as it was done in the case of the neural network models in 
Section \ref{sss:hyp_tune}.

%% file: Sections/05a_Simulation_results.tex
\section{Results}
\label{sec:results}


In this section, the results of different elements of the proposed data-driven DRL-based control learning pipeline are presented. First, the evaluations of the room temperature and bidirectional EV (dis-)charging models are shown and analysed. Then, the simulation results of applying the DRL algorithm to the room temperature control are illustrated, followed by the  results on the joint control of the room temperature and EV (dis-)charging operations. 
Finally, the experimental results demonstrating the DRL agent applied to the real building are presented.


\subsection{Simulation results}

\subsubsection{Evaluation of the EV battery model}

\begin{figure}
  \centering
  \includegraphics[width=\columnwidth]{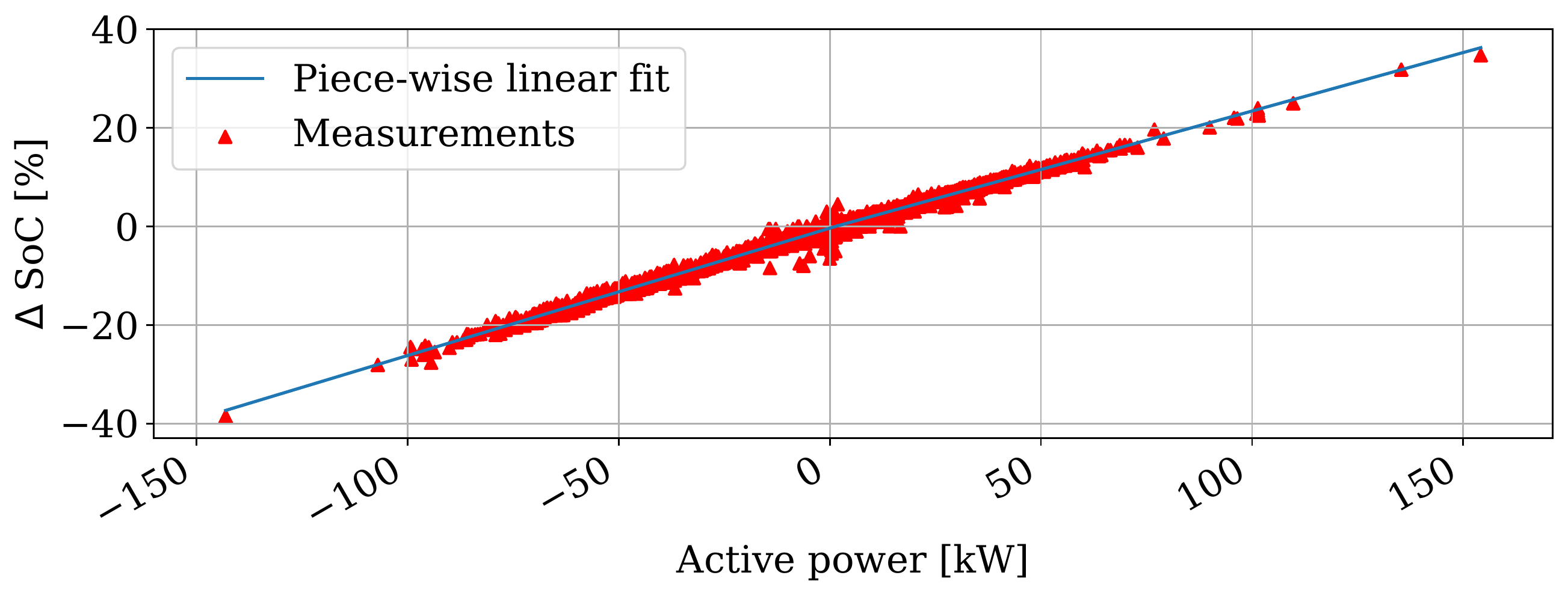}
  \caption{Piece-wise linear EV battery charging/discharging model.}
  \label{fig:bat_model_fit}
\end{figure}

The piece-wise linear EV battery charging/discharging model, together with the real data collected at NEST used for fitting, can be
seen in Fig. \ref{fig:bat_model_fit}.

\begin{figure}
  \centering
    {{\includegraphics[width=\columnwidth]{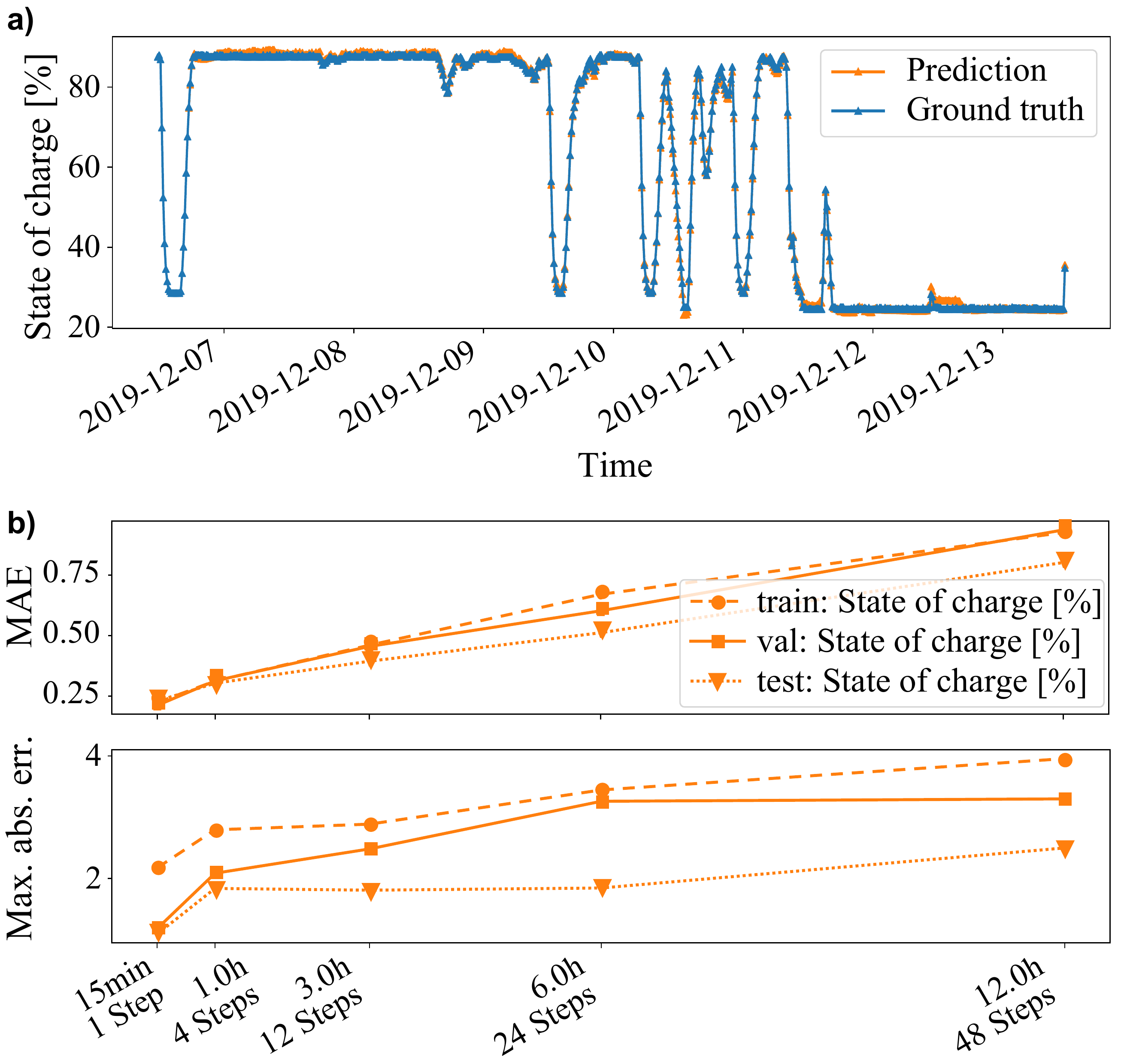}}}%
    \caption{Prediction performance of the EV battery model.  
    a) A sample week from the training set. b) Quantitative evaluation.}%
    \label{fig:bat_6h_eval}%
\end{figure}

The 6h ahead SoC prediction using the EV battery model described 
in Section \ref{ssec:bat} is shown in Fig. \ref{fig:bat_6h_eval}a. 
Note that the ground truth is shown for comparison and was not used to fit the model.
We also performed a more detailed analysis of the prediction performance of the battery model by analysing the mean absolute error (MAE) and maximum absolute error for a different number of prediction steps, up to \SI{12}{\hour} prediction interval (Fig. \ref{fig:bat_6h_eval}b).
The prediction captures the dynamics very well, with an MAE of the SoC of less than 0.75 \% after 6h.   On average, after \SI{12}{\hour}, the prediction will be less than $1\%$ away from the true SoC.  


\subsubsection{Evaluation of the weather model}

We compare two methods for the weather model: a linear model and a recurrent neural network model. As a linear model, we chose a $5$-fold cross-validated
multi-task Lasso estimator from SKLearn \cite{scikit-learn}. For
the RNN, we used the same configuration as the other RNNs in this study (see Sec. \ref{ss:RNN_architecture}).
Both models used the same inputs to make the predictions, i.e.
data from the previous $19$ steps. Further, we used clipping at $0$
for the irradiance in both cases for a fair comparison. Note that 
this makes the model previously described as 
linear actually only piece-wise linear. 

    

    

\begin{figure}
  \centering
  \includegraphics[page=1,width=\columnwidth]
  {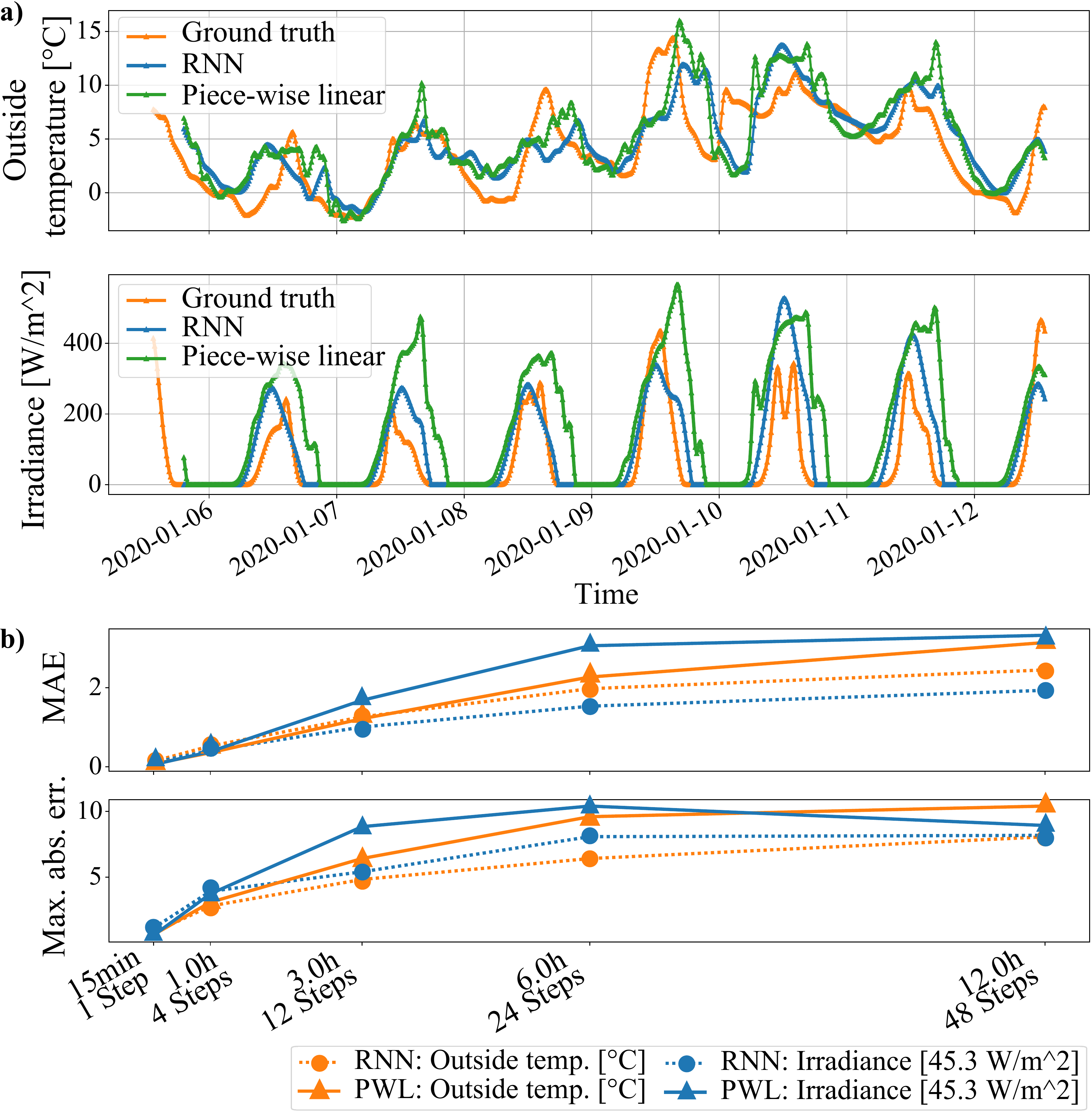}
  \caption{Weather models: 
  a) Prediction over several days - example. b) Quantitative comparison.}
  \label{fig:weather_errors}
\end{figure}

Fig. \ref{fig:weather_errors} shows how the weather
model performs when evaluated
on the test set for one specific initial condition. It can be observed that
the piece-wise linear model makes smoother predictions and diverges faster than the RNN model.
The quality of predictions drops with the longer horizon and, overall, the RNNs provide better predictions, even though the linear model is comparable on short horizons. 

Note that, by investing more thoughts into the piece-wise linear model, 
e.g. by manual feature engineering, one might obtain a 
linear model that may be able to outperform the RNN. 
On the other hand, as the dataset
grows with time, it is easy to increase the 
size of the RNN to make it more powerful, 
which is not the case for the linear model, which is another reason the RNN was 
favoured.

\subsubsection{Evaluation of room temperature model}
\begin{figure}
    \centering
    
    
    {\includegraphics[width=\columnwidth]
     {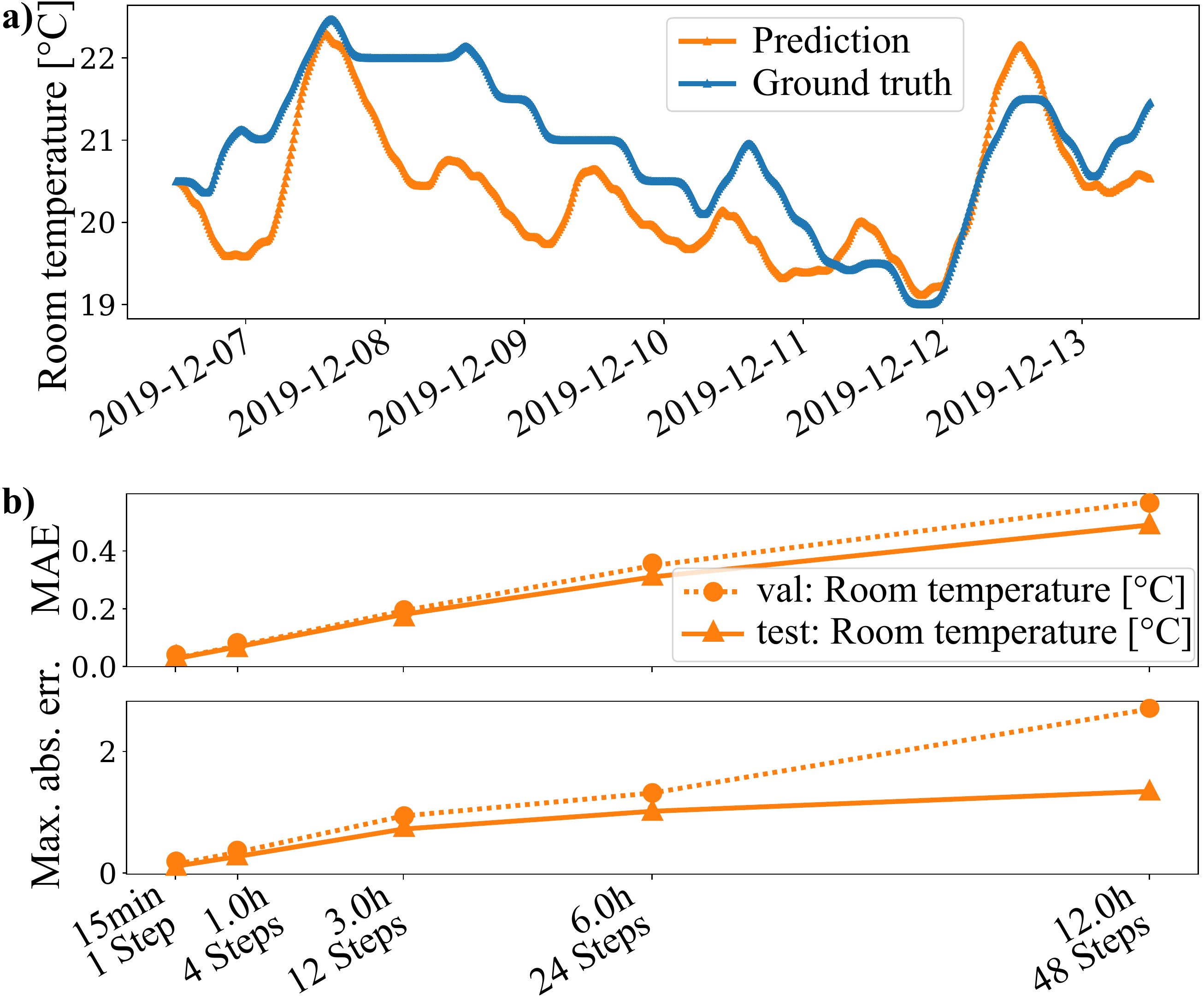}}
    \caption{Room temperature prediction models: Comparison of the piece-wise linear model and the 
    RNN as trained for the weather model. a) Qualitative comparison - example.  b) Quantitative evaluation.}
    \label{fig:room_temp_1week}%
\end{figure}

The performance of the room temperature model is shown in Fig. \ref{fig:room_temp_1week}. 
A quantitative evaluation of the model is shown in Fig. \ref{fig:room_temp_1week}b, where the temperature
prediction is done over a whole week. The MAE and maximum absolute errors are \SI{0.5}{\celsius} and \SI{2.3}{\celsius} after 12h, respectively. As this RNN model showed a satisfactory fit, we selected it as an environment to 
train the DRL agent. 

Note that the quality of the room model influences the final control performance. One known issue is that  black-box models, i.e. non-physics based models, do not extrapolate well for unseen data. In our case, the room temperature model could be outputting physically inconsistent behaviours in the worst case. For example, on a winter day with low solar irradiance and the heating turned off, a black-box model might predict an increase of the room temperature. Such inconsistent physical outputs of the room temperature model can influence the control policy search negatively, as the DRL agent could learn that it could heat the room by closing the heating valves. Therefore, the more phys\-i\-cal\-ly-consistent behaviour a room model expresses for the test data, the better control performance of the DRL agent is expected. However, a detailed analysis of the physical inconsistency of the room temperature model for some input data is outside of the scope of this work.


\subsubsection{Evaluation of the DRL agent for room temperature control}
\label{sss:heat_ag}

We evaluated the DRL agent for both heating and cooling seasons, either by taking two different agents, one for each season, or by letting a unique agent learn the global control policy.
We tested both approaches and obtained better results for the separate agents. 
The reasons for better results in the case of heating only or cooling only agent is that it makes the problem less complex. 
In that way, the deep learning (DDPG) agent is able to find a better control policy. 

\begin{figure*}
\centering
      \includegraphics[page=1,width=0.75\textwidth]{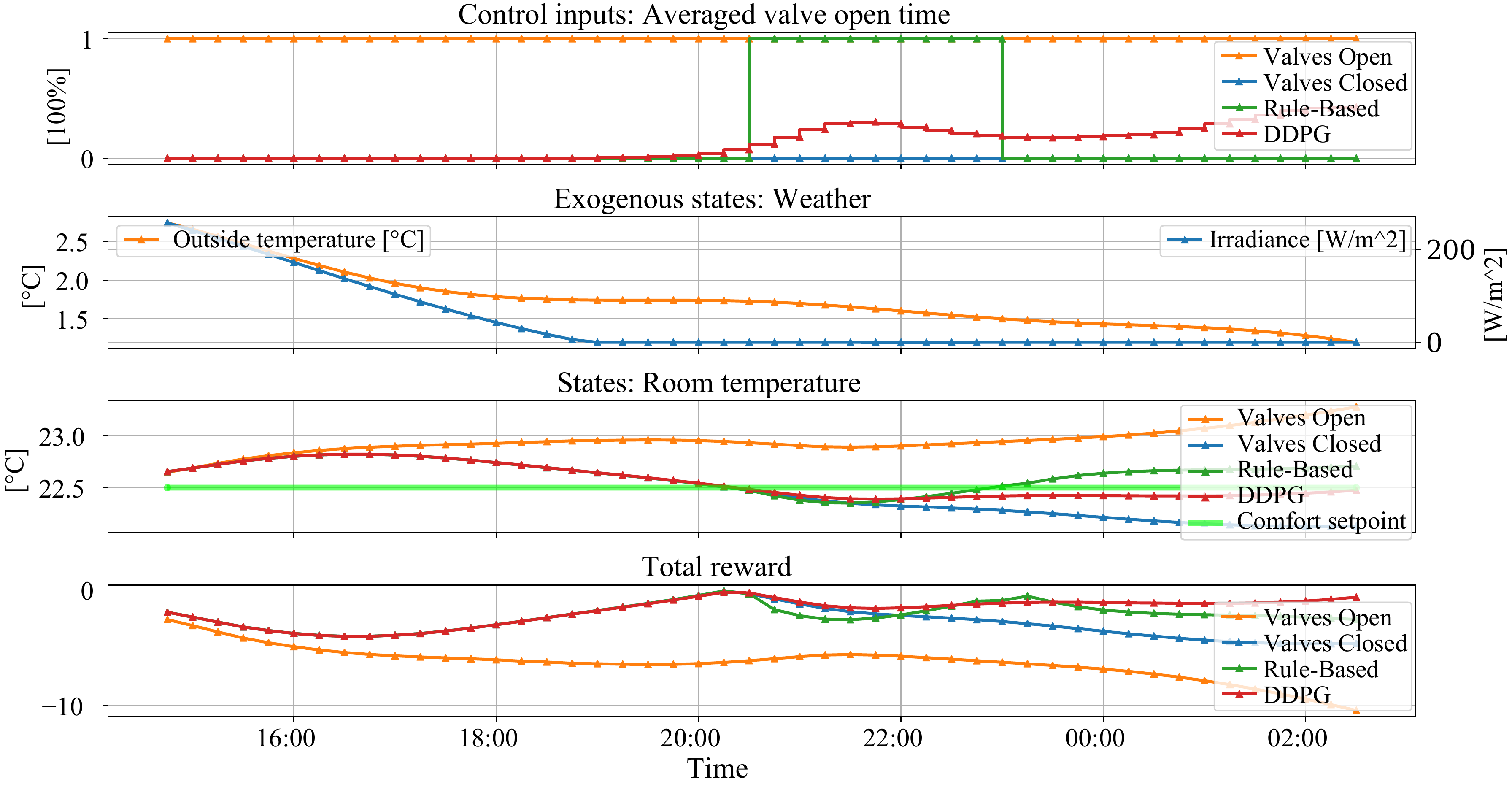}
      \caption{\centering Heating case demonstration. The inflow water temperature was \SI{30.4}{\celsius}.}
      \label{fig:heating_case_demonstration}
\end{figure*}

\begin{figure}
\centering
    \includegraphics[page=1,width=\columnwidth]{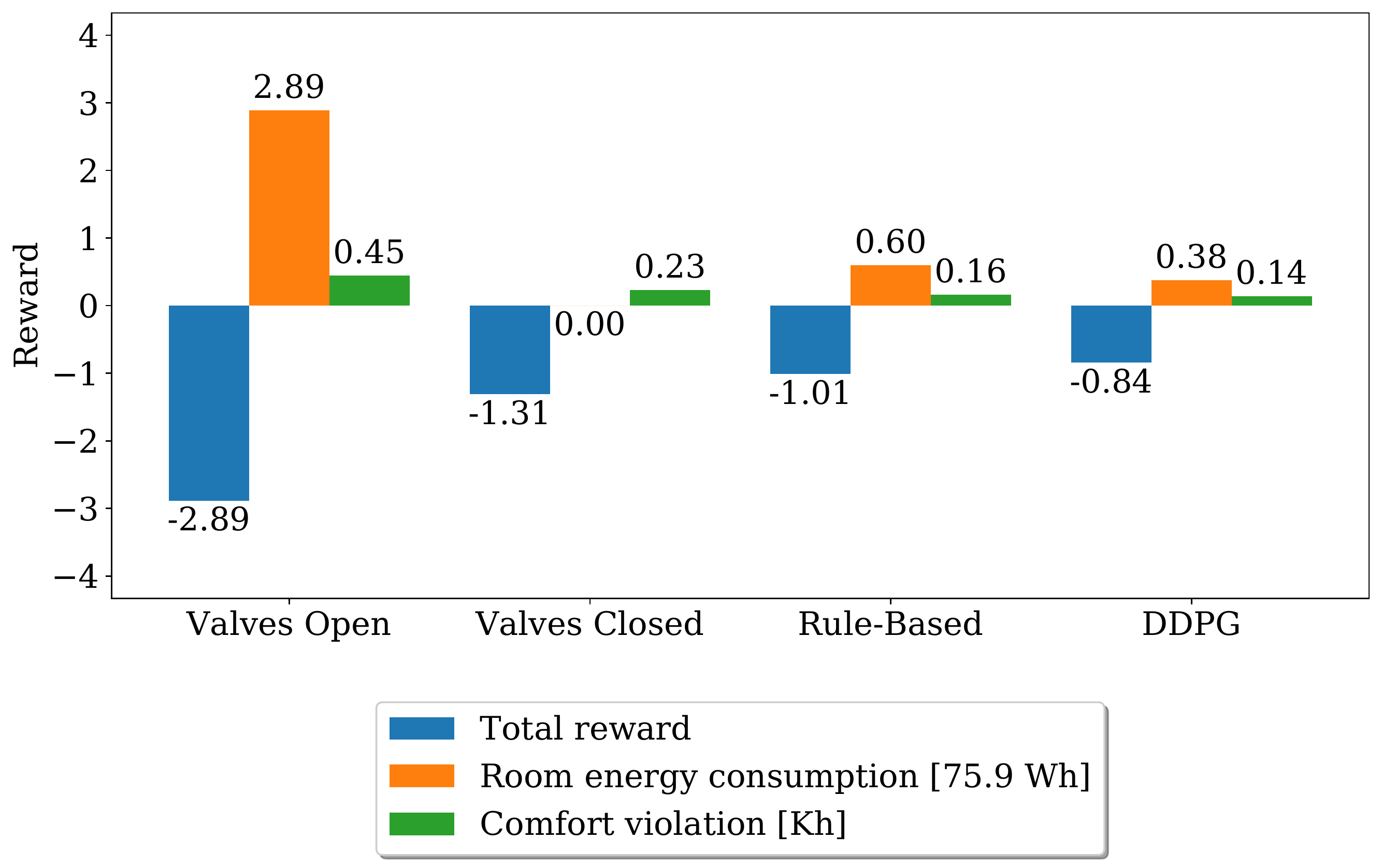}
    \caption{Heating case rewards - example. Quantitative analysis of the results in Fig. \ref{fig:heating_case_demonstration}.}
      \label{fig:heat_eval_ts}
\end{figure}

\begin{figure}
  \centering
  \includegraphics[page=1,width=\columnwidth]{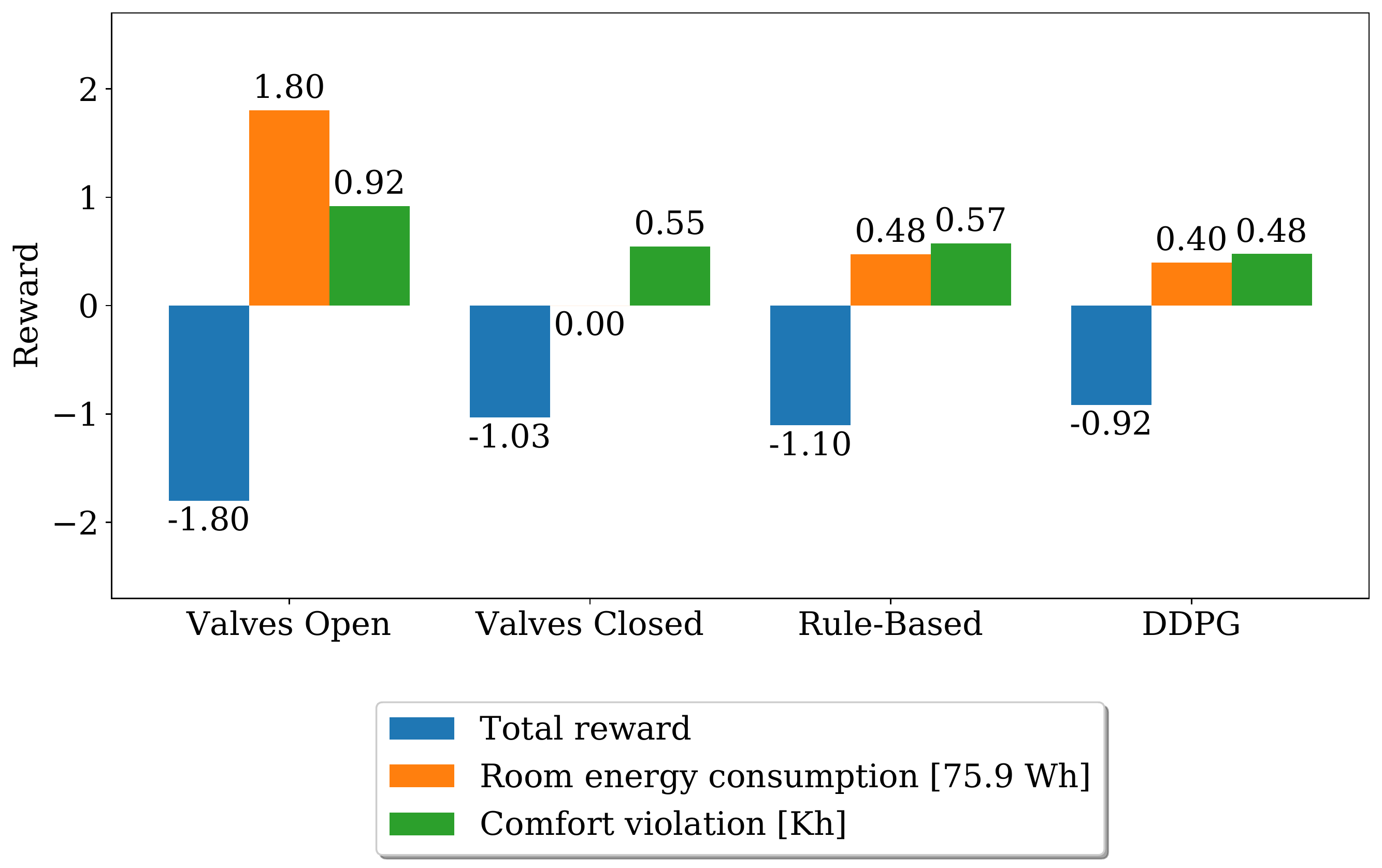}
  \caption{Heating agent evaluation -- average over the whole heating season.}
  \label{fig:heat_eval}
\end{figure}

It actually turned out that for heating cases only, the optimization of the DDPG agent was much harder than in the case of searching for a global control policy and required some manual hyperparameter tuning to perform well. 
Therefore, we decided to switch to a reference  tracking mode by setting $r_{max} = r_{min} = \SI{22.5}{\celsius}$. This makes it easier for the agent to know what actions are beneficial for temperature control since the temperature 
bound violation is only exactly zero for 
$r_t = r_{max} = \SI{22.5}{\celsius}$.
As soon as $r_t$  differs, the comfort violation will increase and the agent is penalized.
We trained the RL agent for $20'000$ steps and the evaluation is shown in Fig. \ref{fig:heat_eval}, where the agent is compared to the following three controllers: one always opening the valves, one always closing them, and a rule-based bang-bang controller without hysteresis, which is a standard industrial controller. 
One can observe that the DDPG agent achieves on average 17\% energy savings and 19\% better comfort satisfaction  compared to the rule-based controller.

A simulated case example is shown in Fig. \ref{fig:heating_case_demonstration}. The DDPG agent can accurately control the room temperature by starting to open the valve before the RB controller, i.e.  before the temperature reaches the setpoint, and opening them only a little to avoid overshooting. 
One can observe that the DDPG agent obtained the least comfort violations while using less energy than the rule-based agent. 
The quantitative analysis of this example shows 36\% energy saving and 13\% better comfort (see Fig. \ref{fig:heat_eval_ts}).

\subsubsection{Evaluation of the joint room heating and EV charging control}

As in the previous case of room temperature control, we again use three controllers as a comparison for the evaluation:

\begin{itemize}
    \item \textbf{Valves Open, Charge:} This agent always leaves the valves open, as the \textit{Valves Open} agent in the previous setting, but additionally always charges the battery at full power instantaneously upon arrival of the EV until it is full.
    \item \textbf{Valves Closed, Discharge:} This agent does the opposite of the previous one, i.e. it never opens the valves and constantly tries to discharge the battery at full power.
    \item \textbf{Rule-Based:} This agent does the same as the previous Rule-Based agent for the heating and constantly charges the battery at full power.
\end{itemize}

\begin{figure}
  \centering
  \includegraphics[page=1,width=\columnwidth]{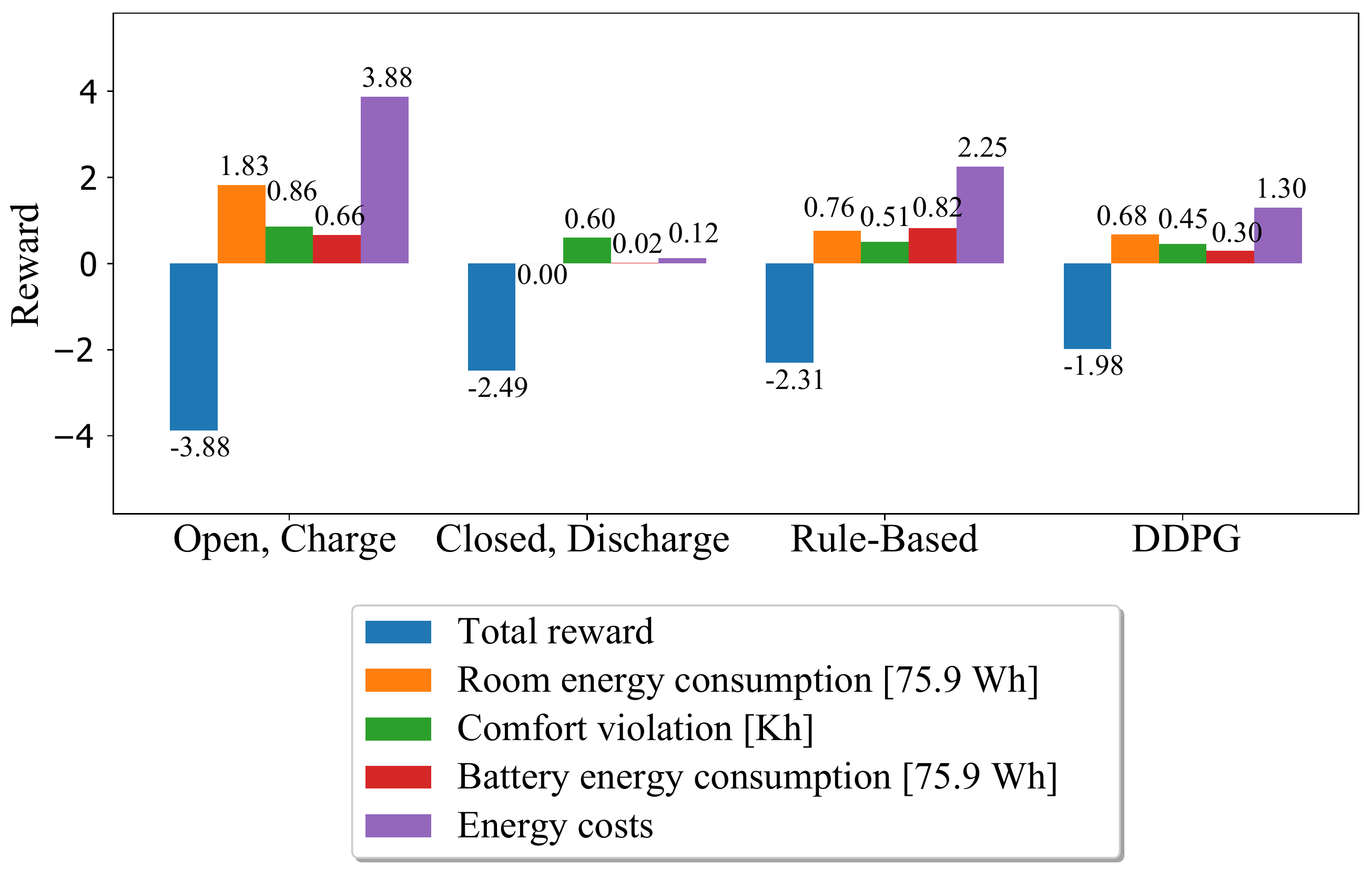}
  \caption{Joint EV charging and room heating control agent evaluation over a 
  total of 10'000 steps. }
  \label{fig:bat_eval}
\end{figure}

The performance of a MIMO DDPG agent
trained on the joint environment is shown in Fig.  \ref{fig:bat_eval}. For the room temperature control, we used the same parameters as in Section 
\ref{sss:heat_ag} and we  considered only heating cases. While again being able
to reduce the comfort violations and the heating energy usage compared to the 
RB agent, the DDPG agent also achieved lower costs. As  expected, the agent
that never turns the heating on and discharges the battery uses the least energy, which also resulted in the lowest costs. Additionally, comfort violations are less pronounced than in the case of constant heating (constantly valve kept on), but still worse than in both RB and DDPG controlled cases. 

\begin{figure*}
\centering
      \includegraphics[page=1,width=0.9\textwidth]{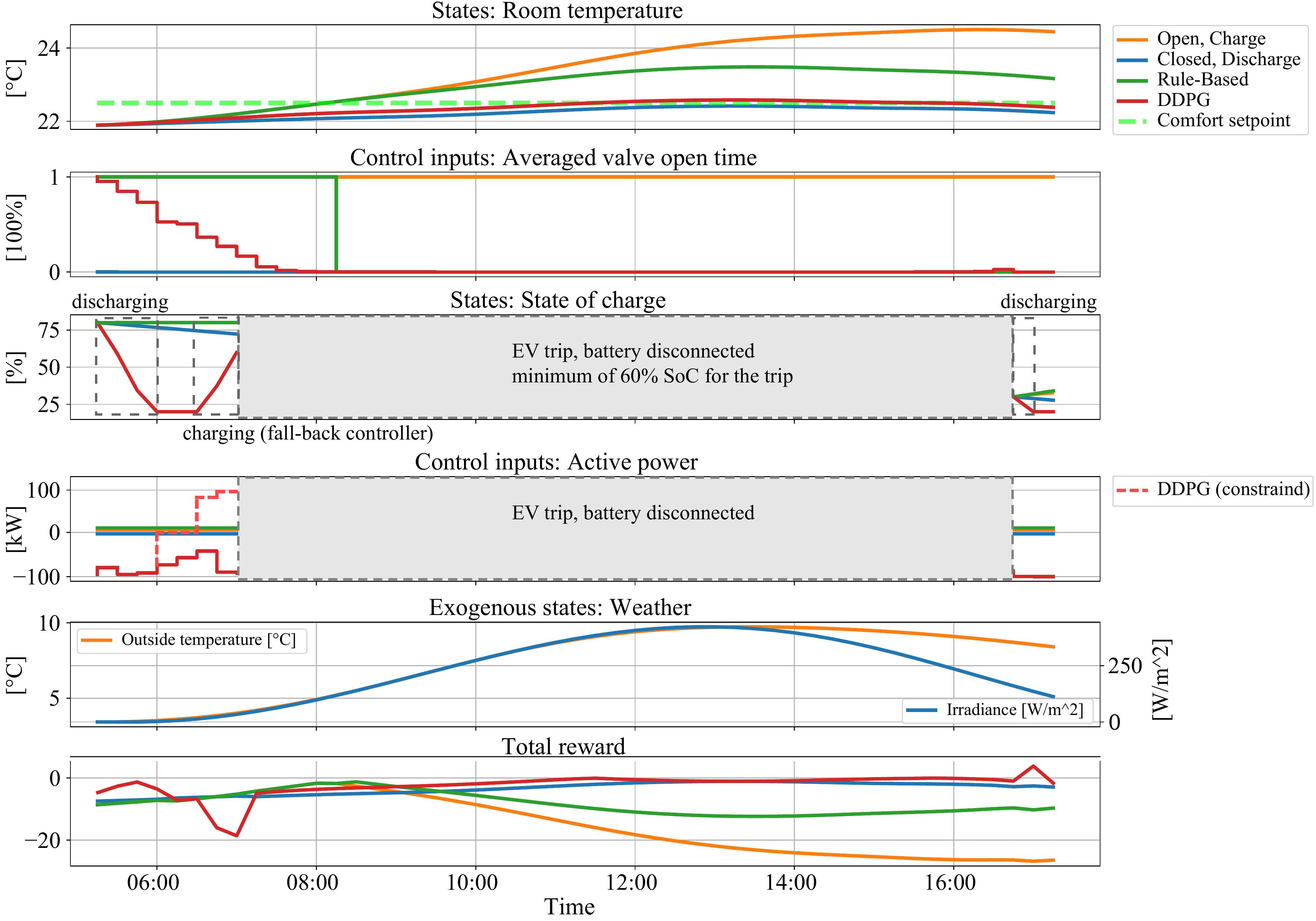}
      
      \caption{Joint room heating control and EV charging. Evaluation of control agents in simulation - example. Inflow water temperature: \SI{29.4}{\celsius}.}
      \label{fig:JointControl_time_series}
\end{figure*}

\begin{figure}
  \centering
  \includegraphics[page=1,width=\columnwidth]{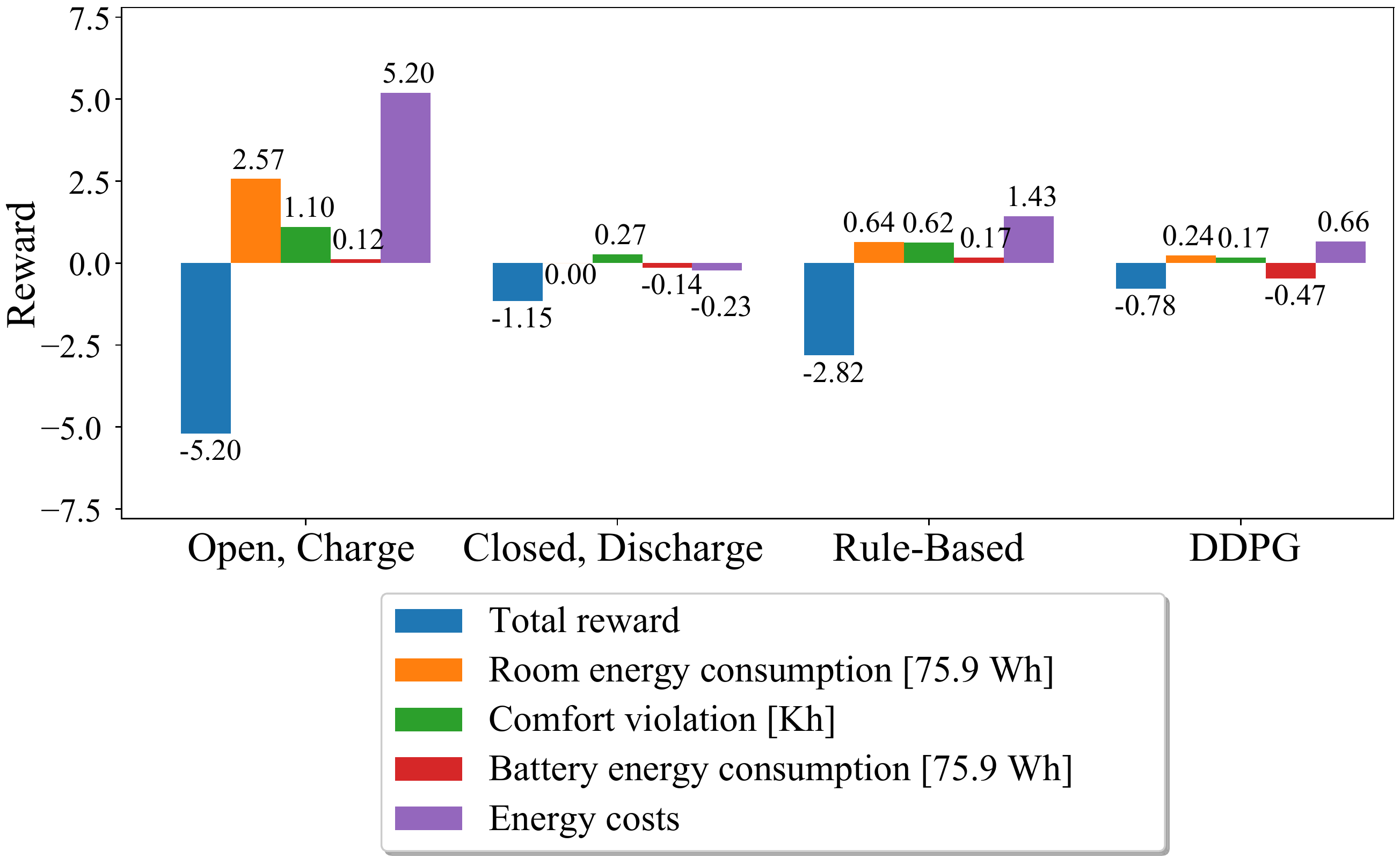}
  \caption{\centering Joint EV charging and room heating control agent evaluation - Statistics of the example plotted in Figure \ref{fig:JointControl_time_series}. }
  \label{fig:JointControl_example_statistics}
\end{figure}

A simulated example is shown in Fig. \ref{fig:JointControl_time_series}. The DDPG agent manages to  regulate the comfort better by using the energy stored in the EV battery. Compared to the RB controller for heating, which heats at the maximum power while the temperature is lower than the reference temperature of \SI{22.5}{\celsius}, the  DDPG controller actively regulates the valves so that better tracking is achieved. In terms of EV battery management, the energy from the EV battery is immediately used at the beginning of the interval until the minimum level of 20\% of SoC is reached, which makes sense due to the lower electricity tariff at this time. 
Then, before the start of the next trip, the fallback controller charges the EV battery to the required SoC. The DDPG control output is presented in full red line, while the  constrained DDPG is shown in the dashed light red line. The quantitative analysis of this DDPG agent is shown in Fig. \ref{fig:JointControl_example_statistics}, where it achieves 63\% energy savings, 71\% better comfort, and 53\% costs savings compared to two RB controllers, for a certain weighting factor between the energy cost savings and comfort satisfaction. Note that this result is specific to the weighting factor used in the reward function. 
On average, when tested over 10'000 historical intervals, the MIMO DDPG controller achieved 12\% better comfort satisfaction, 11\% energy savings, 63\% less EV charging at home, and 42\% energy costs savings compared to two standard RB controllers, for the same weighting factor.


%% file: Sections/05b_Experimental_results.tex
\subsection{Experimental results}

\begin{figure*}
\centering
      \includegraphics[page=1,width=1\textwidth]{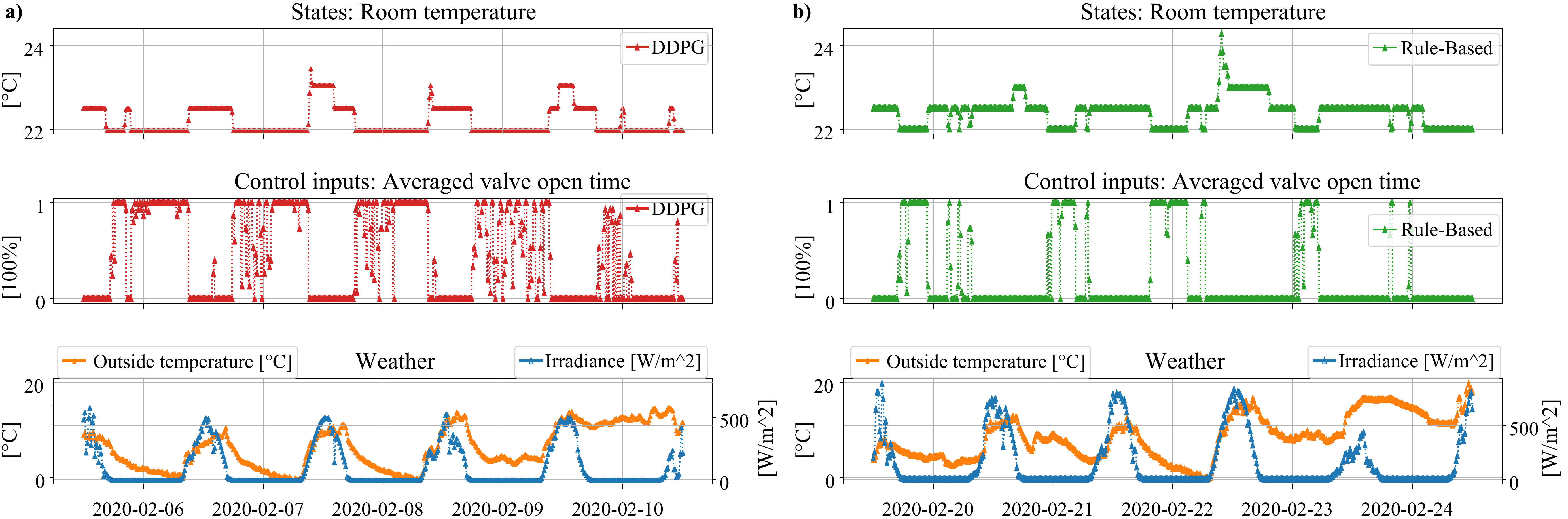}
      \caption{Comparison of DRL (DDPG) control agent a) and classical, RB controller b) at the DFAB HOUSE room 471 at Empa, Duebendorf in Switzerland.
      }
      \label{fig:ExpComparison-time_series}
\end{figure*}

\begin{figure}
  \centering
  \includegraphics[page=1,width=1\columnwidth]{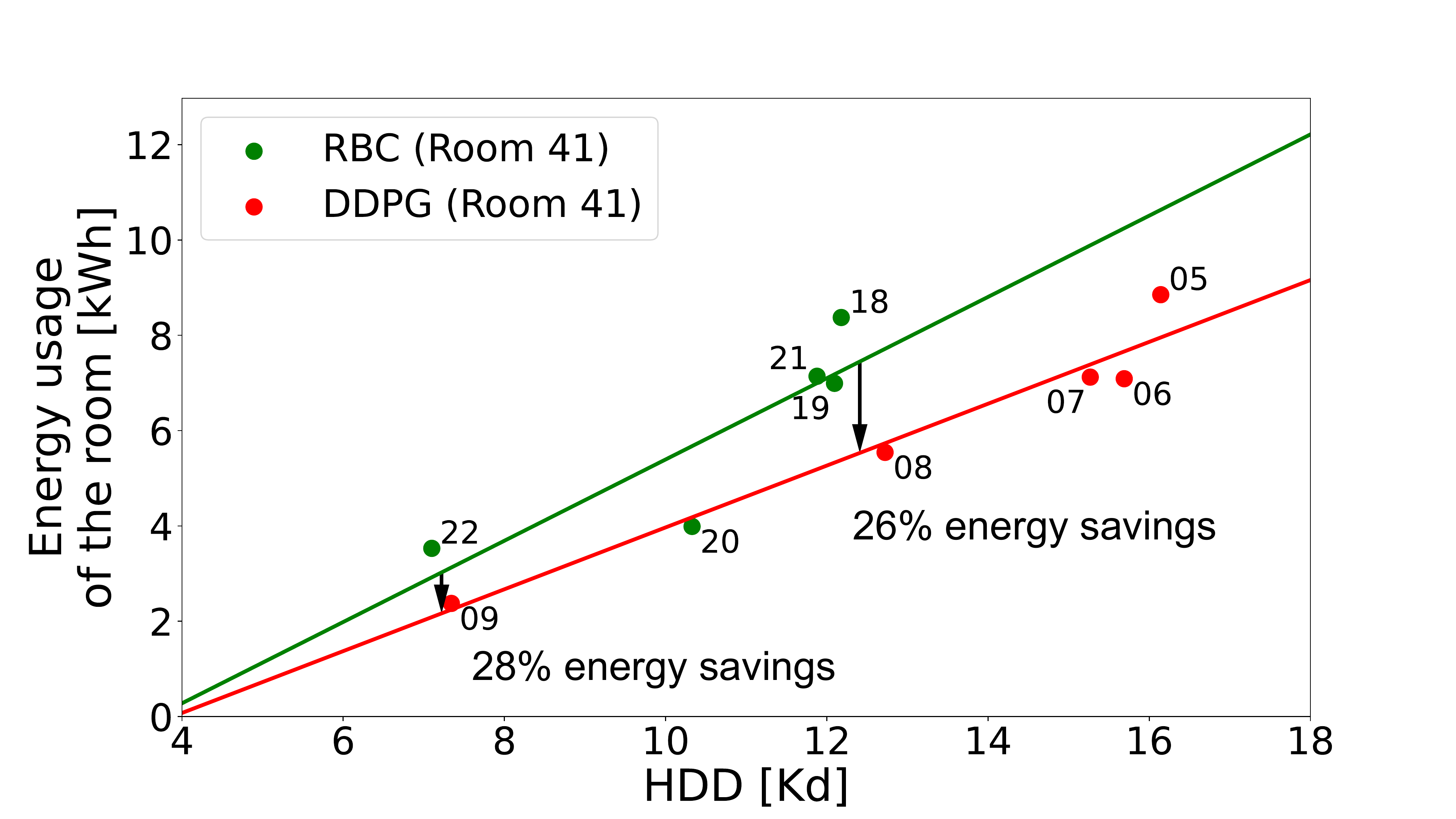}
  \caption{Experimental comparison of the DDPG and rule-based controller performances in term of required energy over the heating degree days. The DDPG provides in average 27\% energy saving and better comfort at the same time. The numbers in the plot correspond to days in February 2020.}
  \label{fig:hdd}
\end{figure}

The DRL control agent, which was obtained in Section \ref{sss:heat_ag} for the heating season and tested in simulation,  was applied on the real building, the DFAB HOUSE, in room 471, for two weeks in February 2020. The performance of the DRL controller was compared with the performance of the room temperature bang-bang RB controller implemented in the same room over a subsequent week. 
The time-series results are shown in Fig. \ref{fig:ExpComparison-time_series}.  Both controllers are aiming at the setpoint  \SI{22.5}{\celsius}. Due to the chosen weighting factor emphasizing energy savings, the DDPG controller is using less energy, at the cost of comfort, keeping the temperature slightly under the setpoint ($-$\SI{0.3}{\celsius} on average). On the other hand, the RBC is staying closer to the setpoint ($-$\SI{0.1}{\celsius} on average), but it is using more energy. 

As the ambient conditions were naturally different for both controllers, we compared them
using the Heating Degree Days (HDD) as a normalization variable. As per definition, the HDD of a given day represents how far from \SI{18}{\celsius} the daily average temperature is \cite{de2016practical}. In other words, higher heating degree days mean lower average outside temperature, for which we naturally expect more energy to be needed. The outside temperature was indeed approximately \SI{4}{\celsius}  lower during the DDPG experiment, which forced the controller to use more energy and made it hard to compare both experiments without a normalization procedure. 

The daily energy used by both the DDPG and the RBC during five experimental days each are plotted against the corresponding HDD in Fig. \ref{fig:hdd}. We can see that the DDPG controller outperforms the RBC:  at HDD levels of around 7 and 12.5, the energy savings are 28\% and 26\%, respectively. On the other hand, we can also observe that while both controllers used between 6 and 8 kWh during three days, the average outside temperature was much lower (about 4\textdegree{}C colder) during the DDPG experiment. In other words, the DDPG algorithm was able to use the same energy budget and maintain similar comfort levels to the RB approach but in harsher conditions.

Additionally, the points in Figure \ref{fig:hdd} exhibit linear-like behaviours. To leverage that fact, we fitted a linear regression to both controllers to capture their global behaviour. This allowed us to clearly picture the gap between the RB algorithm and our proposed method, which on average saves around 25-30\% energy.

%% file: Sections/06_Conslusions_and_discussion.tex
\section{Conclusion and Discussion}
\label{sec:conclusion_and_Discussion}

In this paper, we introduced a fully data-driven DRL-based method to obtain optimal control policies for MIMO building control problems. We demonstrated the method on the joint control of room temperature and bidirectional EV \linebreak (dis-)charging 
to minimise the energy consumption and maximise occupants thermal comfort while ensuring enough energy stored in the EV upon leaving for the next trip. 
We demonstrated the method on a real building case study -- the DFAB HOUSE at Empa Duebendorf in 
Switzerland with available past operational data of less than a year. 


We show in simulation that the trained DRL agents are capable of saving on average 17\% energy over the whole heating season while providing 19\% better comfort satisfaction compared to a classical rule-based controller. When an EV is additionally connected to the building and two tariff electricity pricing is considered, the DRL agents can successfully leverage its battery and decrease the overall cost of electricity. The obtained DRL control agent achieved 12\% better comfort satisfaction, 11\% energy savings, and 42\% energy costs savings 
compared to two standard RB controllers, one controlling the room temperature and another controlling the bidirectional EV (dis-)charging. This result is specific to the weighting factor used in the DRL algorithm to balance the energy cost savings and comfort satisfaction.
Finally, we demonstrate a successful transfer of the learnt DRL policy from simulation onto the actual building achieving up to 30\% energy savings while maintaining similar comfort compared to a conventional RB room temperature controller over three weeks during the heating season. 

The data-driven DRL-based control method proposed in this work  
is a viable approach to satisfy all  the \textit{BA industry requirements} for control of modern buildings, as defined in Section  \ref{sec:BA_ind_reqs}. We demonstrated that this method could match the first three {BA industry requirements}. In terms of the fourth requirement on transferability (and usability) for similar control problems in other buildings, we can argue in favour of the developed method; This  method is suitable for use on any other building to obtain a room temperature controller or a joint (MIMO) control of room temperature and bidirectional EV charging. One can reuse the same NN and DRL architectures structure 
to obtain the control policies. 

We applied the same methodology to another room at the DFAB HOUSE, and we obtained comparable results. We believe that this method has a strong potential to work for any building or room, and could thus be a stepping stone towards obtaining transferable model-free data-driven room temperature control policies.
As such, we also believe it to be valuable for the BA industry 
due to its potential for transferability, as it minimises the engineering efforts to obtain a custom-tailored controller for each room and building of interest while optimizing the energy savings and occupant comfort satisfaction.

However, we still need to address a few points before this method can achieve widespread transferability to any building or room. 

The availability and quality of the building model is the first point to be addressed and explored. As demonstrated in this paper, 
the building model could be built as an RNN model, which could be directly applied to another room with the same setting, i.e. the same HVAC equipment and the same number of sensors and actuators. However, rooms generally differ in terms of HVAC equipment and the number of sensors and actuators. Thus, to model a different room, a certain engineering effort needs to be invested into linking the new inputs and outputs to the RNN model and fitting it. This process could be simplified and even automated if a linked, i.e. semantic, database of a building exists. 

Secondly, the availability of past building operational data is a requirement to apply our black-box pipeline. While this may not be an issue for existing building with operational data stored in databases, it could be an issue for new or retrofitted buildings. A potential solution to this could be to apply transfer learning to the modelling part of this method and learn the dynamics of a new building with fewer data. 
Similarly, transfer learning could be applied to "jump-start" the learning of the control policy for another building, given already existing proven policies in other buildings. This is also directly related to the last \textit{BA industry requirement} 
on (self-) adaptability and continuous commissioning of building controllers. Transferring a controller to another, unseen building or re-applying it to the same building after a retrofit or whenever a change of dynamics is observed, is in essence very similar problem and a very interesting direction for future work.

%% file: Sections/10_Acknowledgement.tex
\section*{Acknowledgements}
    
This research was funded by Empa Duebendorf in Switzerland.

\section*{Declaration of competing interests}
    
The authors declare that they have no known competing financial interests or personal relationships that could have appeared to influence the work reported in this paper.

%% file: Sections/20_Appendix.tex

\section{Hyperparameters of the RNN}

The hyperparameters of the RNN model of the room temperature dynamics are shown in Table \ref{tab:hyp_rnn}.

\section{SoC constraint satisfaction}
\label{app:bat_env}

Assume we have the piece-wise linear battery model as described
in section \ref{ssec:bat}, omitting the $bat$ superscript and
using $p = p_{t+1}$ for clarity:
\begin{equation}
    \begin{split}
    m^{bat}: \mathcal{S} \times \mathcal{P} &\rightarrow \mathcal{S}\\
    (s_t, p) &\mapsto \hat{s}_{t + 1} := m^{bat}(s_t, p) 
    \end{split}
\end{equation}

with:
\begin{equation}
    \begin{split}
    \hat{s}_{t + 1} &= m^{bat}(s_t, p) \\
            &= s_t + \Delta s(p)\\
            &= s_t + \alpha_0 + \alpha_1 p + \alpha_2 \max \{0, p\}
    \end{split}
\end{equation}

We assume that the coefficients $\alpha_i$ have the
following properties to make sure the model can be inverted and
 is physically meaningful.

\begin{itemize}
    \item $\alpha_1 > 0$: The more the battery is discharged 
        (i.e. the more negative the active power), 
        the faster the SoC decreases.
    \item $-\alpha_1 < \alpha_2 < 0$: The slope of $\Delta s(p)$
        is always positive, but smaller for charging (for $p > 0$). I.e. one
        cannot gain energy from charging and discharging again. 
    \item $\alpha_0 \leq 0$, small: The battery does not increase its SoC
        when not used.
\end{itemize}

Using these properties, we will define next how to make
sure that bounds on the SoC will be satisfied.

\begin{table}
    \centering
    \begin{tabular}{
    |p{0.07\columnwidth}
    |p{0.35\columnwidth}
    |p{0.07\columnwidth}
    |p{0.14\columnwidth}
    |p{0.14\columnwidth}|}
        \hline
        Sym\-bol & Meaning & Dom\-ain & Room temp. model & Weath\-er model\\
        \hline\hline
        $n_l$ & Number of recurrent layers & $\mathbb{N}$ & 3 & 1\\
        $n_c$ & Number of neurons per layer & $\mathbb{N}$ & 30 & 60 \\
        $n_{ep}$ & Number of epochs in training & $\mathbb{N}$ & 10 & 80\\
        $\eta$ & Learning rate & $\mathbb{R}^+$ 
        & $1.544\mathrm{e}{-5}$ & $6.163\mathrm{e}{-5}$
        \\
        $\sigma_i$ & Standard deviation of the Gaussian noise added to the input & $\mathbb{R}^+$ & $3.633\mathrm{e}{-6}$
        & 0.01202\\
        $Cell$ & Type of recurrent cell used &
        & \textit{LSTM} & \textit{GRU}\\
    \hline
    \end{tabular}
    \caption{Hyperparameters of the RNN}
    \label{tab:hyp_rnn}
\end{table}

\subsection{Minimum and maximum SoC constraints}

We require the SoC of the battery to lie within predefined bounds $[s_{min}, s_{max}]$ at any time. Assuming we start
from $s_t$, it suffices to show that the next SoC, $s_{t+1}$, given the previous SoC, stays within the bounds, and then apply the argument
recursively. 
For the maximum constraint, 
we have to make sure that:
\begin{equation}
    \begin{split}
    s_{t + 1} = s_t + \alpha_0 + \alpha_1 p + \alpha_2 \max \{0, p\} &\leq s_{max} \\
    \Leftrightarrow \alpha_1 p + \alpha_2 \max \{0, p\} &\leq s_{max} - s_t - \alpha_0\\
    \end{split}
\end{equation}

Let us define the following helper function:
\begin{equation}
     h(p) :=
  \begin{cases}
        \alpha_1 + \alpha_2 & p > 0 \\
        \alpha_1 & \text{else} \\
  \end{cases}
\end{equation}

Note that it is positive for all values of $p$
because of the properties on the coefficients $\alpha_i$.
Now we can rewrite the equation above as:
\begin{equation}
    \begin{split}
    p \cdot h(p)    &\leq s_{max} - s_t - \alpha_0\\
    \Leftrightarrow p    &\leq \frac{s_{max} - s_t - \alpha_0}{h(p)}\\
    \end{split}
\end{equation}

To get a bound for $p$ from this equation that does not
contain $p$ itself, we need to make a case distinction:
\begin{itemize}
    \item \textbf{Case 1: $s_{max} - s_t - \alpha_0 > 0$}
        This means that the SoC at the next step will be lower than the maximum SoC
        when $p=0$, therefore we can discharge as much as we want, i.e.
        we do not need to handle the case $p<0$, so we only look at $p > 0$,
        therefore we have $h(p)=\alpha_1 + \alpha_2$
        
    \item \textbf{Case 2: $s_{max} - s_t - \alpha_0 < 0$}
        This means that the SoC at the next step will be higher than maximum SoC
        when $p=0$, therefore we need to discharge in any case, i.e.
        $p < 0$, which means $h(p) = \alpha_1$
\end{itemize}

Putting the two cases together, we get the
following bound on the active power $p$:
\begin{equation}
  \label{eq:s_max}
     p \leq c^{bat}(s_{max}, s_t; \alpha_i) :=
  \begin{cases}
        \frac{s_{max} - s_t - \alpha_0}{\alpha_1 + \alpha_2} & s_{max} - s_t - \alpha_0 > 0 \\
        \frac{s_{max} - s_t - \alpha_0}{\alpha_1} & \text{else} \\
  \end{cases}
\end{equation}

Note that in the edge case $s_{max} - s_t - \alpha_0 = 0$
both cases return the same, i.e. the bound is continuous.
Applying the same chain of reasoning to the $s_{min}$ case, one
can derive the following: 
\begin{equation}
    \label{eq:s_min}
    p \geq c^{bat}(s_{min}, s_t; \alpha_i)
\end{equation}

Note that this case is using the exact same function.

\subsection{Achieving the goal SoC}

We want to ensure that the battery is  charged
for some minimum desired amount $s_{des}$ at a given time $t_{des}$.
Assuming we are now at time $t$, i.e. the SoC is $s_t$, and
assuming we can charge for a maximum of $p_{max}$, then at the next timestep, the SoC has to be at least 
$s_{des} - (t_{des} - t - 1) \cdot \Delta s(p_{max})$, where $(t_{des} - t - 1)$
denotes the number of remaining timesteps after this step.
Now we can use the results derived in the previous section from the equation \ref{eq:s_min} and 
get:
\begin{equation}
    \label{eq:s_des}
    p \geq p_{min}^{des} := c^{bat}(s_{des} - (t_{des} - t - 1) \cdot \Delta s(p_{max}), s_t; \alpha_i)
\end{equation}

Note that, if we start with an SoC that is already too low to achieve the
goal SoC, the bounds will require an active power $p > p_{max}$,
which is not possible, and $p_{max}$ would be applied.

\subsection{Constraining battery controller}
\label{appendix_constraining_bat_control}

Now we can finally combine all the previous constraints
to define the controller that constrains the active power for
the battery charging and discharging. We consider the following constraints:

\begin{itemize}
    \item Direct constraints: $p_{min} \leq p \leq p_{max}$
    \item SoC constraints: $s_{min} \leq s_t \leq s_{max}$
    \item Charging constraint: $s_t \geq s_{des}$ for $t = t_{des}$
\end{itemize}

Note that we still use $p := p_{t+1}$. Using the formulas defined 
before, the last two constraints can be converted to constraints on $p$
as shown in equations \eqref{eq:s_max}, \eqref{eq:s_min} and \eqref{eq:s_des}.
Combining these constraints with the direct constraints on $p$ and 
choosing always the tightest one yields:
\begin{equation}
    \label{eq:p_con}
    \begin{split}
        p \geq \tilde{p}_{min} &:= \max \{ p_{min}, p_{min}^{des},
            c^{bat}(s_{min}, s_t; \alpha_i)  \} \\
        p \leq \tilde{p}_{max} &:= \min \{ p_{max}, c^{bat}(s_{max}, s_t; \alpha_i) \} \\
    \end{split}
\end{equation}

Finally, we can define our safety controller that assures that
the chosen action, i.e. the active power, lies in the appropriate range.
\begin{equation}
    \label{eq:f_safe}
    f^{safe}(p) := clip_{[ \tilde{p}_{min}, \tilde{p}_{max} ] }(p)
\end{equation}

where $clip$ is the clipping function defined as follows:
\begin{equation}
    clip_{[a, b]}(p) := 
        \begin{cases}
            a & p \leq a \\
            b & p \geq b \\
            p &  \text{else} \\
        \end{cases}
\end{equation}

Note that the function $f^{safe}(p)$ implicitly depends on a lot of
parameters, i.e. $p_{min}, p_{max}, s_{min}, \allowbreak s_{max}, 
s_{des}, s_t, \allowbreak t, t_{des}$ and the parameters of the model $\alpha_i$
and not only on $p$.

\section{Data preparation}
\label{sec:data_proc}

\subsection{DFAB data}

The following variables are measured inside the DFAB unit and
are processed as follows, before their usage in the data-driven
learning process.
\begin{itemize}
\item \textbf{Room temperature ($r_t$):} The room temperature contained
a few data points at exactly \SI{0}{\celsius} which were removed.
Furthermore, sequences of constant temperature that
lasted for at least one day were removed, too. In the next step, 
spikes in the temperature of a magnitude of at least
\SI{1.5}{\celsius} were extracted and deleted. Finally, we applied
Gaussian smoothing with a standard deviation of $5.0$.
\item \textbf{Valves ($u_t$):} The data measured for each individual valve
only stated if the valve was open (1) or closed (0). Subsampling
resulted in values in the interval $[0.0, 1.0]$. Since this series
naturally contains long sequences of constant values, i.e. $0$ or $1$,
we only removed constant sequences which lasted for at least $30$ days.
\item \textbf{Water temperatures ($h_t^{in}, h_t^{out}$):} 
The water temperature of the heating water
flowing into and out of the rooms was processed by removing all
data points that did not lie in the range 
$[\SI{10.0}{\celsius}, \SI{50.0}{\celsius}]$ were removed
and then smoothing with a Gaussian filter with a standard
deviation of $5.0$ was applied.
\end{itemize}

\subsection{Weather data}

Outside temperature and solar irradiance are measured by 
the weather station at NEST. They were processed
in the following way.
\begin{itemize}
\item \textbf{Outside temperature ($o_t$):} First, we remove values that
are constant for more than $30$ minutes. In the next step,
we fill values that are missing by linear
interpolation between the last and the next known value, but only
if the time interval of missing values was less than 
$45$ minutes. Finally, we smooth the data with a
Gaussian filter with a standard deviation of $2.0$.
\item \textbf{Irradiance ($i_t$):} Since the irradiance data series
naturally contains values that are constant for a long time,
e.g. zero at night, we only remove a series of data points
if they are constant for at least \SI{20}{\hour}. Then again 
we fill missing data points by interpolation and 
smooth the data as was done with the temperature data.
\end{itemize}

\subsection{EV battery data}
\label{ssec:bat_filt}

The data of the battery consists of
the state of charge (SoC) and the active power used to charge
or discharge the battery. The two time series were
processed as follows.
\begin{itemize}
\item \textbf{State of charge ($s^{bat}_t$):} 
Since the SoC cannot lie outside of the interval
$[0.0\%, 100.0\%]$, we remove all values that lie outside that 
range including the boundary values. Further, if the data
is exactly constant for at least \SI{24}{\hour}, we assume something went wrong
with the data collection and remove the data of that time
interval.
\item \textbf{Active power ($p_t$):} In this case, we do not have strict
boundaries for the values, so we only remove values 
where the series was constant for at least 6h.
\end{itemize}

\section{Implementation}
\label{sec:impl}

The work was implemented in Python version $3.6.6$ and is not compatible to versions
$3.5$ and lower since f-strings were used. The main libraries that were used are listed in Table 
\ref{tab:python_libs}. Note that the most recent 
version of all libraries was used,
except for TensorFlow \cite{tensorflow2015} because
of a dependency on another library, Keras-RL \cite{plappert2016kerasrl}.
In most cases, the produced code is 
Pep-8. 
The actual code can be accessed at
\url{https://github.com/chbauman/MasterThesis}. There is also
information available on how to run the code. 

\begin{table}
    \centering
    \begin{tabular}{
    |p{0.7\columnwidth}
    |p{0.2\columnwidth}|}
        \hline
        Library & Version\\
        \hline\hline
        Numpy \cite{numpy} & 1.18.1 \\
        TensorFlow \cite{tensorflow2015} & 1.14.0 \\
        Keras \cite{chollet2015keras} & 2.3.1 \\
        Hyperopt \cite{bergstra2015hyperopt} & 0.2.3 \\
        Pandas \cite{mckinney2011pandas} & 0.25.3 \\
        SkLearn \cite{scikit-learn} & 0.22.1 \\
        Matplotlib \cite{mpl} & 3.1.2 \\
        OpenAI gym \cite{1606.01540} & 0.15.4 \\
        Keras-RL \cite{plappert2016kerasrl} & 0.4.2 \\
        SciPy \cite{2020SciPy} & 1.4.1 \\
        Statsmodels \cite{seabold2010statsmodels} & 0.10.2 \\
    \hline
    \end{tabular}
    \caption{Python libraries, used with Python \cite{python} version 3.6.6}
    \label{tab:python_libs}
\end{table}

\subsection{Data whitening}

As another data processing step, we whitened the data, i.e. it was scaled
to have mean $0.0$ and variance $1.0$ before training
the models. This is a standard procedure in machine
learning and helps to avoid a bias in the feature importance while
also allowing task-independent weight initialization in the neural
network training. Since this was done manually, without the use of
an existing library, this resulted in a few complications. For example, 
the reinforcement learning environment took the original actions as
input and then had to scale them, feed them to the model and 
scale the output of the model back to the original domain to get
the output for the agent.

%% file: MAIN.bbl
\begin{thebibliography}{86}
\expandafter\ifx\csname natexlab\endcsname\relax\def\natexlab#1{#1}\fi
\providecommand{\bibinfo}[2]{#2}
\ifx\xfnm\relax \def\xfnm[#1]{\unskip,\space#1}\fi
\bibitem[{Change et~al.(2014)}]{change2014mitigation}
\bibinfo{author}{I.~C. Change}, et~al.,
\newblock \bibinfo{title}{Mitigation of climate change},
\newblock \bibinfo{journal}{Contribution of Working Group III to the Fifth
  Assessment Report of the Intergovernmental Panel on Climate Change}
  \bibinfo{volume}{1454} (\bibinfo{year}{2014}).
  \bibinfo{note}{\url{https://www.ipcc.ch/report/ar5/wg3/}}.
\bibitem[{Ramesh et~al.(2010)Ramesh, Prakash, and Shukla}]{ramesh2010life}
\bibinfo{author}{T.~Ramesh}, \bibinfo{author}{R.~Prakash},
  \bibinfo{author}{K.~Shukla},
\newblock \bibinfo{title}{Life cycle energy analysis of buildings: An
  overview},
\newblock \bibinfo{journal}{Energy and buildings} \bibinfo{volume}{42}
  (\bibinfo{year}{2010}) \bibinfo{pages}{1592--1600}.
  \bibinfo{note}{\url{https://doi.org/10.1016/j.enbuild.2010.05.007}}.
\bibitem[{Shaikh et~al.(2014)Shaikh, Nor, Nallagownden, Elamvazuthi, and
  Ibrahim}]{shaikh2014review}
\bibinfo{author}{P.~H. Shaikh}, \bibinfo{author}{N.~B.~M. Nor},
  \bibinfo{author}{P.~Nallagownden}, \bibinfo{author}{I.~Elamvazuthi},
  \bibinfo{author}{T.~Ibrahim},
\newblock \bibinfo{title}{A review on optimized control systems for building
  energy and comfort management of smart sustainable buildings},
\newblock \bibinfo{journal}{Renewable and Sustainable Energy Reviews}
  \bibinfo{volume}{34} (\bibinfo{year}{2014}) \bibinfo{pages}{409--429}.
  \bibinfo{note}{\url{https://doi.org/10.1016/j.rser.2014.03.027}}.
\bibitem[{Huang et~al.(2020)Huang, Lovati, Zhang, and
  Bales}]{huang2020coordinated}
\bibinfo{author}{P.~Huang}, \bibinfo{author}{M.~Lovati},
  \bibinfo{author}{X.~Zhang}, \bibinfo{author}{C.~Bales},
\newblock \bibinfo{title}{A coordinated control to improve performance for a
  building cluster with energy storage, electric vehicles, and energy sharing
  considered},
\newblock \bibinfo{journal}{Applied Energy} \bibinfo{volume}{268}
  (\bibinfo{year}{2020}) \bibinfo{pages}{114983}.
  \bibinfo{note}{\url{https://doi.org/10.1016/j.apenergy.2020.114983}}.
\bibitem[{Chel and Kaushik(2018)}]{chel2018renewable}
\bibinfo{author}{A.~Chel}, \bibinfo{author}{G.~Kaushik},
\newblock \bibinfo{title}{Renewable energy technologies for sustainable
  development of energy efficient building},
\newblock \bibinfo{journal}{Alexandria Engineering Journal}
  \bibinfo{volume}{57} (\bibinfo{year}{2018}) \bibinfo{pages}{655--669}.
  \bibinfo{note}{\url{https://doi.org/10.1016/j.aej.2017.02.027}}.
\bibitem[{Chwieduk(2003)}]{chwieduk2003towards}
\bibinfo{author}{D.~Chwieduk},
\newblock \bibinfo{title}{Towards sustainable-energy buildings},
\newblock \bibinfo{journal}{Applied energy} \bibinfo{volume}{76}
  (\bibinfo{year}{2003}) \bibinfo{pages}{211--217}.
  \bibinfo{note}{\url{https://doi.org/10.1016/S0306-2619(03)00059-X}}.
\bibitem[{Zhou et~al.(2019)Zhou, Cao, Hensen, and Lund}]{zhou2019energy}
\bibinfo{author}{Y.~Zhou}, \bibinfo{author}{S.~Cao}, \bibinfo{author}{J.~L.
  Hensen}, \bibinfo{author}{P.~D. Lund},
\newblock \bibinfo{title}{Energy integration and interaction between buildings
  and vehicles: A state-of-the-art review},
\newblock \bibinfo{journal}{Renewable and Sustainable Energy Reviews}
  \bibinfo{volume}{114} (\bibinfo{year}{2019}) \bibinfo{pages}{109337}.
  \bibinfo{note}{\url{https://doi.org/10.1016/j.rser.2019.109337}}.
\bibitem[{Liu et~al.(2013)Liu, Chau, Wu, and Gao}]{liu2013opportunities}
\bibinfo{author}{C.~Liu}, \bibinfo{author}{K.~Chau}, \bibinfo{author}{D.~Wu},
  \bibinfo{author}{S.~Gao},
\newblock \bibinfo{title}{Opportunities and challenges of vehicle-to-home,
  vehicle-to-vehicle, and vehicle-to-grid technologies},
\newblock \bibinfo{journal}{Proceedings of the IEEE} \bibinfo{volume}{101}
  (\bibinfo{year}{2013}) \bibinfo{pages}{2409--2427}.
  \bibinfo{note}{\url{https://doi.org/10.1109/JPROC.2013.2271951}}.
\bibitem[{Park and Nagy(2018)}]{park2018comprehensive}
\bibinfo{author}{J.~Y. Park}, \bibinfo{author}{Z.~Nagy},
\newblock \bibinfo{title}{Comprehensive analysis of the relationship between
  thermal comfort and building control research-a data-driven literature
  review},
\newblock \bibinfo{journal}{Renewable and Sustainable Energy Reviews}
  \bibinfo{volume}{82} (\bibinfo{year}{2018}) \bibinfo{pages}{2664--2679}.
  \bibinfo{note}{\url{https://doi.org/10.1016/j.rser.2017.09.102}}.
\bibitem[{Salsbury(2005)}]{salsbury2005survey}
\bibinfo{author}{T.~I. Salsbury},
\newblock \bibinfo{title}{A survey of control technologies in the building
  automation industry},
\newblock \bibinfo{journal}{IFAC Proceedings Volumes} \bibinfo{volume}{38}
  (\bibinfo{year}{2005}) \bibinfo{pages}{90--100}.
\bibitem[{Verhelst et~al.(2017)Verhelst, Van~Ham, Saelens, and
  Helsen}]{verhelst2017model}
\bibinfo{author}{J.~Verhelst}, \bibinfo{author}{G.~Van~Ham},
  \bibinfo{author}{D.~Saelens}, \bibinfo{author}{L.~Helsen},
\newblock \bibinfo{title}{Model selection for continuous commissioning of
  hvac-systems in office buildings: A review},
\newblock \bibinfo{journal}{Renewable and Sustainable Energy Reviews}
  \bibinfo{volume}{76} (\bibinfo{year}{2017}) \bibinfo{pages}{673--686}.
  \bibinfo{note}{\url{https://doi.org/10.1016/j.rser.2017.01.119}}.
\bibitem[{Stluka et~al.(2018)Stluka, Parthasarathy, Gabel, and
  Samad}]{stluka2018architectures}
\bibinfo{author}{P.~Stluka}, \bibinfo{author}{G.~Parthasarathy},
  \bibinfo{author}{S.~Gabel}, \bibinfo{author}{T.~Samad},
\newblock \bibinfo{title}{Architectures and algorithms for building
  automation—an industry view},
\newblock in: \bibinfo{booktitle}{Intelligent Building Control Systems},
  \bibinfo{publisher}{Springer}, \bibinfo{year}{2018}, pp.
  \bibinfo{pages}{11--43}.
  \bibinfo{note}{\url{https://doi.org/10.1007/978-3-319-68462-8_2}}.
\bibitem[{Samad et~al.(2020)Samad, Bauer, Bortoff, Di~Cairano, Fagiano,
  Odgaard, Rhinehart, S{\'a}nchez-Pe{\~n}a, Serbezov, Ankersen
  et~al.}]{samad2020industry}
\bibinfo{author}{T.~Samad}, \bibinfo{author}{M.~Bauer},
  \bibinfo{author}{S.~Bortoff}, \bibinfo{author}{S.~Di~Cairano},
  \bibinfo{author}{L.~Fagiano}, \bibinfo{author}{P.~F. Odgaard},
  \bibinfo{author}{R.~R. Rhinehart}, \bibinfo{author}{R.~S{\'a}nchez-Pe{\~n}a},
  \bibinfo{author}{A.~Serbezov}, \bibinfo{author}{F.~Ankersen}, et~al.,
\newblock \bibinfo{title}{Industry engagement with control research:
  Perspective and messages},
\newblock \bibinfo{journal}{Annual Reviews in Control} \bibinfo{volume}{49}
  (\bibinfo{year}{2020}) \bibinfo{pages}{1--14}.
  \bibinfo{note}{\url{https://doi.org/10.1016/j.arcontrol.2020.03.002}}.
\bibitem[{Skogestad and Postlethwaite(2007)}]{skogestad2007multivariable}
\bibinfo{author}{S.~Skogestad}, \bibinfo{author}{I.~Postlethwaite},
  \bibinfo{title}{Multivariable feedback control: analysis and design},
  volume~\bibinfo{volume}{2}, \bibinfo{publisher}{Citeseer},
  \bibinfo{year}{2007}.
  \bibinfo{note}{\url{https://dl.acm.org/doi/abs/10.5555/525126}}.
\bibitem[{Privara et~al.(2013)Privara, Cigler, V{\'a}{\v{n}}a, Oldewurtel,
  Sagerschnig, and {\v{Z}}{\'a}{\v{c}}ekov{\'a}}]{privara2013building}
\bibinfo{author}{S.~Privara}, \bibinfo{author}{J.~Cigler},
  \bibinfo{author}{Z.~V{\'a}{\v{n}}a}, \bibinfo{author}{F.~Oldewurtel},
  \bibinfo{author}{C.~Sagerschnig},
  \bibinfo{author}{E.~{\v{Z}}{\'a}{\v{c}}ekov{\'a}},
\newblock \bibinfo{title}{Building modeling as a crucial part for building
  predictive control},
\newblock \bibinfo{journal}{Energy and Buildings} \bibinfo{volume}{56}
  (\bibinfo{year}{2013}) \bibinfo{pages}{8--22}.
  \bibinfo{note}{\url{https://doi.org/10.1016/j.enbuild.2012.10.024}}.
\bibitem[{Jain et~al.(2018)Jain, Nghiem, Morari, and
  Mangharam}]{jain2018learning}
\bibinfo{author}{A.~Jain}, \bibinfo{author}{T.~Nghiem},
  \bibinfo{author}{M.~Morari}, \bibinfo{author}{R.~Mangharam},
\newblock \bibinfo{title}{Learning and control using gaussian processes},
\newblock in: \bibinfo{booktitle}{2018 ACM/IEEE 9th International Conference on
  Cyber-Physical Systems (ICCPS)}, \bibinfo{organization}{IEEE}, pp.
  \bibinfo{pages}{140--149}.
  \bibinfo{note}{\url{https://doi.org/10.1109/ICCPS.2018.00022}}.
\bibitem[{Serale et~al.(2018)Serale, Fiorentini, Capozzoli, Bernardini, and
  Bemporad}]{serale2018model}
\bibinfo{author}{G.~Serale}, \bibinfo{author}{M.~Fiorentini},
  \bibinfo{author}{A.~Capozzoli}, \bibinfo{author}{D.~Bernardini},
  \bibinfo{author}{A.~Bemporad},
\newblock \bibinfo{title}{Model predictive control (mpc) for enhancing building
  and hvac system energy efficiency: Problem formulation, applications and
  opportunities},
\newblock \bibinfo{journal}{Energies} \bibinfo{volume}{11}
  (\bibinfo{year}{2018}) \bibinfo{pages}{631}.
  \bibinfo{note}{\url{https://doi.org/10.3390/en11030631}}.
\bibitem[{Oldewurtel et~al.(2012)Oldewurtel, Parisio, Jones, Gyalistras,
  Gwerder, Stauch, Lehmann, and Morari}]{oldewurtel2012use}
\bibinfo{author}{F.~Oldewurtel}, \bibinfo{author}{A.~Parisio},
  \bibinfo{author}{C.~N. Jones}, \bibinfo{author}{D.~Gyalistras},
  \bibinfo{author}{M.~Gwerder}, \bibinfo{author}{V.~Stauch},
  \bibinfo{author}{B.~Lehmann}, \bibinfo{author}{M.~Morari},
\newblock \bibinfo{title}{Use of model predictive control and weather forecasts
  for energy efficient building climate control},
\newblock \bibinfo{journal}{Energy and Buildings} \bibinfo{volume}{45}
  (\bibinfo{year}{2012}) \bibinfo{pages}{15--27}.
  \bibinfo{note}{\url{https://doi.org/10.1016/j.enbuild.2011.09.022}}.
\bibitem[{Xu et~al.(2010)Xu, Wang, and Huang}]{xu2010robust}
\bibinfo{author}{X.~Xu}, \bibinfo{author}{S.~Wang}, \bibinfo{author}{G.~Huang},
\newblock \bibinfo{title}{Robust mpc for temperature control of
  air-conditioning systems concerning on constraints and multitype
  uncertainties},
\newblock \bibinfo{journal}{Building Services Engineering Research and
  Technology} \bibinfo{volume}{31} (\bibinfo{year}{2010})
  \bibinfo{pages}{39--55}.
  \bibinfo{note}{\url{https://doi.org/10.1177/0143624409352420}}.
\bibitem[{Tanaskovic et~al.(2017)Tanaskovic, Sturzenegger, Smith, and
  Morari}]{tanaskovic2017robust}
\bibinfo{author}{M.~Tanaskovic}, \bibinfo{author}{D.~Sturzenegger},
  \bibinfo{author}{R.~Smith}, \bibinfo{author}{M.~Morari},
\newblock \bibinfo{title}{Robust adaptive model predictive building climate
  control},
\newblock \bibinfo{journal}{Ifac-Papersonline} \bibinfo{volume}{50}
  (\bibinfo{year}{2017}) \bibinfo{pages}{1871--1876}.
  \bibinfo{note}{\url{https://doi.org/10.1016/j.ifacol.2017.08.257}}.
\bibitem[{Khosravi et~al.(2019)Khosravi, Schmid, Eichler, Heer, and
  Smith}]{khosravi2019machine}
\bibinfo{author}{M.~Khosravi}, \bibinfo{author}{N.~Schmid},
  \bibinfo{author}{A.~Eichler}, \bibinfo{author}{P.~Heer},
  \bibinfo{author}{R.~S. Smith},
\newblock \bibinfo{title}{Machine learning-based modeling and controller tuning
  of a heat pump},
\newblock in: \bibinfo{booktitle}{Journal of Physics: Conference Series},
  volume \bibinfo{volume}{1343}, \bibinfo{organization}{IOP Publishing}, p.
  \bibinfo{pages}{012065}.
  \bibinfo{note}{\url{https://doi.org/10.1088/1742-6596/1343/1/012065}}.
\bibitem[{Khosravi et~al.(2021)Khosravi, Behrunani, Myszkorowski, Smith,
  Rupenyan, and Lygeros}]{khosravi2021performance}
\bibinfo{author}{M.~Khosravi}, \bibinfo{author}{V.~Behrunani},
  \bibinfo{author}{P.~Myszkorowski}, \bibinfo{author}{R.~S. Smith},
  \bibinfo{author}{A.~Rupenyan}, \bibinfo{author}{J.~Lygeros},
\newblock \bibinfo{title}{Performance-driven cascade controller tuning with
  bayesian optimization},
\newblock \bibinfo{journal}{IEEE Transactions on Industrial Electronics}
  (\bibinfo{year}{2021}).
  \bibinfo{note}{\url{https://doi.org/10.1109/TIE.2021.3050356}}.
\bibitem[{Aswani et~al.(2013)Aswani, Gonzalez, Sastry, and
  Tomlin}]{aswani2013provably}
\bibinfo{author}{A.~Aswani}, \bibinfo{author}{H.~Gonzalez},
  \bibinfo{author}{S.~S. Sastry}, \bibinfo{author}{C.~Tomlin},
\newblock \bibinfo{title}{Provably safe and robust learning-based model
  predictive control},
\newblock \bibinfo{journal}{Automatica} \bibinfo{volume}{49}
  (\bibinfo{year}{2013}) \bibinfo{pages}{1216--1226}.
  \bibinfo{note}{\url{https://doi.org/10.1016/j.automatica.2013.02.003}}.
\bibitem[{Aswani et~al.(2011)Aswani, Master, Taneja, Culler, and
  Tomlin}]{aswani2011reducing}
\bibinfo{author}{A.~Aswani}, \bibinfo{author}{N.~Master},
  \bibinfo{author}{J.~Taneja}, \bibinfo{author}{D.~Culler},
  \bibinfo{author}{C.~Tomlin},
\newblock \bibinfo{title}{Reducing transient and steady state electricity
  consumption in hvac using learning-based model-predictive control},
\newblock \bibinfo{journal}{Proceedings of the IEEE} \bibinfo{volume}{100}
  (\bibinfo{year}{2011}) \bibinfo{pages}{240--253}.
  \bibinfo{note}{\url{https://doi.org/10.1109/JPROC.2011.2161242}}.
\bibitem[{{Chen} et~al.(2018){Chen}, {Shi}, and {Zhang}}]{2018arXiv180511835C}
\bibinfo{author}{Y.~{Chen}}, \bibinfo{author}{Y.~{Shi}},
  \bibinfo{author}{B.~{Zhang}},
\newblock \bibinfo{title}{{Optimal Control Via Neural Networks: A Convex
  Approach}},
\newblock \bibinfo{journal}{arXiv e-prints}  (\bibinfo{year}{2018})
  \bibinfo{pages}{arXiv:1805.11835}.
  \bibinfo{note}{\url{https://ui.adsabs.harvard.edu/abs/2018arXiv180511835C}}.
\bibitem[{Smarra et~al.(2018)Smarra, Jain, de~Rubeis, Ambrosini, D’Innocenzo,
  and Mangharam}]{smarra2018data}
\bibinfo{author}{F.~Smarra}, \bibinfo{author}{A.~Jain},
  \bibinfo{author}{T.~de~Rubeis}, \bibinfo{author}{D.~Ambrosini},
  \bibinfo{author}{A.~D’Innocenzo}, \bibinfo{author}{R.~Mangharam},
\newblock \bibinfo{title}{Data-driven model predictive control using random
  forests for building energy optimization and climate control},
\newblock \bibinfo{journal}{Applied energy} \bibinfo{volume}{226}
  (\bibinfo{year}{2018}) \bibinfo{pages}{1252--1272}.
  \bibinfo{note}{\url{https://doi.org/10.1016/j.apenergy.2018.02.126}}.
\bibitem[{B{\"u}nning et~al.(2020)B{\"u}nning, Huber, Heer, Aboudonia, and
  Lygeros}]{bunning2020experimental}
\bibinfo{author}{F.~B{\"u}nning}, \bibinfo{author}{B.~Huber},
  \bibinfo{author}{P.~Heer}, \bibinfo{author}{A.~Aboudonia},
  \bibinfo{author}{J.~Lygeros},
\newblock \bibinfo{title}{Experimental demonstration of data predictive control
  for energy optimization and thermal comfort in buildings},
\newblock \bibinfo{journal}{Energy and Buildings} \bibinfo{volume}{211}
  (\bibinfo{year}{2020}) \bibinfo{pages}{109792}.
  \bibinfo{note}{\url{https://doi.org/10.1016/j.enbuild.2020.109792}}.
\bibitem[{Sutton and Barto(2018)}]{sutton2018reinforcement}
\bibinfo{author}{R.~S. Sutton}, \bibinfo{author}{A.~G. Barto},
  \bibinfo{title}{Reinforcement learning: An introduction},
  \bibinfo{publisher}{MIT press}, \bibinfo{year}{2018}.
  \bibinfo{note}{\url{https://doi.org/10.1016/S1364-6613(99)01331-5}}.
\bibitem[{LeCun et~al.(2015)LeCun, Bengio, and Hinton}]{lecun2015deep}
\bibinfo{author}{Y.~LeCun}, \bibinfo{author}{Y.~Bengio},
  \bibinfo{author}{G.~Hinton},
\newblock \bibinfo{title}{Deep learning},
\newblock \bibinfo{journal}{nature} \bibinfo{volume}{521}
  (\bibinfo{year}{2015}) \bibinfo{pages}{436--444}.
  \bibinfo{note}{\url{https://doi.org/10.1038/nature14539}}.
\bibitem[{Arulkumaran et~al.(2017)Arulkumaran, Deisenroth, Brundage, and
  Bharath}]{arulkumaran2017deep}
\bibinfo{author}{K.~Arulkumaran}, \bibinfo{author}{M.~P. Deisenroth},
  \bibinfo{author}{M.~Brundage}, \bibinfo{author}{A.~A. Bharath},
\newblock \bibinfo{title}{Deep reinforcement learning: A brief survey},
\newblock \bibinfo{journal}{IEEE Signal Processing Magazine}
  \bibinfo{volume}{34} (\bibinfo{year}{2017}) \bibinfo{pages}{26--38}.
  \bibinfo{note}{\url{https://doi.org/10.1109/MSP.2017.2743240}}.
\bibitem[{Mnih et~al.(2015)Mnih, Kavukcuoglu, Silver, Rusu, Veness, Bellemare,
  Graves, Riedmiller, Fidjeland, Ostrovski et~al.}]{mnih2015human}
\bibinfo{author}{V.~Mnih}, \bibinfo{author}{K.~Kavukcuoglu},
  \bibinfo{author}{D.~Silver}, \bibinfo{author}{A.~A. Rusu},
  \bibinfo{author}{J.~Veness}, \bibinfo{author}{M.~G. Bellemare},
  \bibinfo{author}{A.~Graves}, \bibinfo{author}{M.~Riedmiller},
  \bibinfo{author}{A.~K. Fidjeland}, \bibinfo{author}{G.~Ostrovski}, et~al.,
\newblock \bibinfo{title}{Human-level control through deep reinforcement
  learning},
\newblock \bibinfo{journal}{nature} \bibinfo{volume}{518}
  (\bibinfo{year}{2015}) \bibinfo{pages}{529--533}. \bibinfo{note}{\url
  {http://dx.doi.org/10.1038/nature14236}}.
\bibitem[{Silver et~al.(2017)Silver, Schrittwieser, Simonyan, Antonoglou,
  Huang, Guez, Hubert, Baker, Lai, Bolton et~al.}]{silver2017mastering}
\bibinfo{author}{D.~Silver}, \bibinfo{author}{J.~Schrittwieser},
  \bibinfo{author}{K.~Simonyan}, \bibinfo{author}{I.~Antonoglou},
  \bibinfo{author}{A.~Huang}, \bibinfo{author}{A.~Guez},
  \bibinfo{author}{T.~Hubert}, \bibinfo{author}{L.~Baker},
  \bibinfo{author}{M.~Lai}, \bibinfo{author}{A.~Bolton}, et~al.,
\newblock \bibinfo{title}{Mastering the game of go without human knowledge},
\newblock \bibinfo{journal}{nature} \bibinfo{volume}{550}
  (\bibinfo{year}{2017}) \bibinfo{pages}{354--359}.
  \bibinfo{note}{\url{doi:10.1038/nature24270}}.
\bibitem[{Young et~al.(2018)Young, Hazarika, Poria, and
  Cambria}]{young2018recent}
\bibinfo{author}{T.~Young}, \bibinfo{author}{D.~Hazarika},
  \bibinfo{author}{S.~Poria}, \bibinfo{author}{E.~Cambria},
\newblock \bibinfo{title}{Recent trends in deep learning based natural language
  processing},
\newblock \bibinfo{journal}{ieee Computational intelligenCe magazine}
  \bibinfo{volume}{13} (\bibinfo{year}{2018}) \bibinfo{pages}{55--75}.
  \bibinfo{note}{\url{https://doi.org/10.1109/MCI.2018.2840738}}.
\bibitem[{Esteva et~al.(2019)Esteva, Robicquet, Ramsundar, Kuleshov, DePristo,
  Chou, Cui, Corrado, Thrun, and Dean}]{esteva2019guide}
\bibinfo{author}{A.~Esteva}, \bibinfo{author}{A.~Robicquet},
  \bibinfo{author}{B.~Ramsundar}, \bibinfo{author}{V.~Kuleshov},
  \bibinfo{author}{M.~DePristo}, \bibinfo{author}{K.~Chou},
  \bibinfo{author}{C.~Cui}, \bibinfo{author}{G.~Corrado},
  \bibinfo{author}{S.~Thrun}, \bibinfo{author}{J.~Dean},
\newblock \bibinfo{title}{A guide to deep learning in healthcare},
\newblock \bibinfo{journal}{Nature medicine} \bibinfo{volume}{25}
  (\bibinfo{year}{2019}) \bibinfo{pages}{24--29}.
  \bibinfo{note}{\url{https://doi.org/10.1038/s41591-018-0316-z}}.
\bibitem[{Wang and Hong(2020)}]{wang2020reinforcement}
\bibinfo{author}{Z.~Wang}, \bibinfo{author}{T.~Hong},
\newblock \bibinfo{title}{Reinforcement learning for building controls: The
  opportunities and challenges},
\newblock \bibinfo{journal}{Applied Energy} \bibinfo{volume}{269}
  (\bibinfo{year}{2020}) \bibinfo{pages}{115036}.
  \bibinfo{note}{\url{https://doi.org/10.1016/j.apenergy.2020.115036}}.
\bibitem[{V{\'a}zquez-Canteli and Nagy(2019)}]{vazquez2019reinforcement}
\bibinfo{author}{J.~R. V{\'a}zquez-Canteli}, \bibinfo{author}{Z.~Nagy},
\newblock \bibinfo{title}{Reinforcement learning for demand response: A review
  of algorithms and modeling techniques},
\newblock \bibinfo{journal}{Applied energy} \bibinfo{volume}{235}
  (\bibinfo{year}{2019}) \bibinfo{pages}{1072--1089}. \bibinfo{note}{\url{
  https://doi.org/10.1016/j.apenergy.2018.11.002}}.
\bibitem[{Mason and Grijalva(2019)}]{mason2019review}
\bibinfo{author}{K.~Mason}, \bibinfo{author}{S.~Grijalva},
\newblock \bibinfo{title}{A review of reinforcement learning for autonomous
  building energy management},
\newblock \bibinfo{journal}{Computers \& Electrical Engineering}
  \bibinfo{volume}{78} (\bibinfo{year}{2019}) \bibinfo{pages}{300--312}.
  \bibinfo{note}{\url{https://doi.org/10.1016/j.compeleceng.2019.07.019}}.
\bibitem[{Lillicrap et~al.(2015)Lillicrap, Hunt, Pritzel, Heess, Erez, Tassa,
  Silver, and Wierstra}]{lillicrap2015continuous}
\bibinfo{author}{T.~P. Lillicrap}, \bibinfo{author}{J.~J. Hunt},
  \bibinfo{author}{A.~Pritzel}, \bibinfo{author}{N.~Heess},
  \bibinfo{author}{T.~Erez}, \bibinfo{author}{Y.~Tassa},
  \bibinfo{author}{D.~Silver}, \bibinfo{author}{D.~Wierstra},
\newblock \bibinfo{title}{Continuous control with deep reinforcement learning},
\newblock \bibinfo{journal}{arXiv preprint arXiv:1509.02971}
  (\bibinfo{year}{2015}).
  \bibinfo{note}{\url{https://ui.adsabs.harvard.edu/abs/2015arXiv150902971L}}.
\bibitem[{Ruano et~al.(2006)Ruano, Crispim, Concei{\c{c}}ao, and
  L{\'u}cio}]{ruano2006prediction}
\bibinfo{author}{A.~E. Ruano}, \bibinfo{author}{E.~M. Crispim},
  \bibinfo{author}{E.~Z. Concei{\c{c}}ao}, \bibinfo{author}{M.~M.~J.
  L{\'u}cio},
\newblock \bibinfo{title}{Prediction of building's temperature using neural
  networks models},
\newblock \bibinfo{journal}{Energy and Buildings} \bibinfo{volume}{38}
  (\bibinfo{year}{2006}) \bibinfo{pages}{682--694}.
  \bibinfo{note}{\url{https://doi.org/10.1016/j.enbuild.2005.09.007}}.
\bibitem[{Mustafaraj et~al.(2011)Mustafaraj, Lowry, and
  Chen}]{mustafaraj2011prediction}
\bibinfo{author}{G.~Mustafaraj}, \bibinfo{author}{G.~Lowry},
  \bibinfo{author}{J.~Chen},
\newblock \bibinfo{title}{Prediction of room temperature and relative humidity
  by autoregressive linear and nonlinear neural network models for an open
  office},
\newblock \bibinfo{journal}{Energy and Buildings} \bibinfo{volume}{43}
  (\bibinfo{year}{2011}) \bibinfo{pages}{1452--1460}.
  \bibinfo{note}{\url{https://doi.org/10.1016/j.enbuild.2011.02.007}}.
\bibitem[{Taylor and Stone(2009)}]{taylor2009transfer}
\bibinfo{author}{M.~E. Taylor}, \bibinfo{author}{P.~Stone},
\newblock \bibinfo{title}{Transfer learning for reinforcement learning domains:
  A survey},
\newblock \bibinfo{journal}{Journal of Machine Learning Research}
  \bibinfo{volume}{10} (\bibinfo{year}{2009}).
  \bibinfo{note}{\url{https://www.jmlr.org/papers/volume10/taylor09a/taylor09a.pdf}}.
\bibitem[{Xu et~al.(2020)Xu, Wang, Wang, O'Neill, and Zhu}]{xu2020one}
\bibinfo{author}{S.~Xu}, \bibinfo{author}{Y.~Wang}, \bibinfo{author}{Y.~Wang},
  \bibinfo{author}{Z.~O'Neill}, \bibinfo{author}{Q.~Zhu},
\newblock \bibinfo{title}{One for many: Transfer learning for building hvac
  control},
\newblock in: \bibinfo{booktitle}{Proceedings of the 7th ACM International
  Conference on Systems for Energy-Efficient Buildings, Cities, and
  Transportation}, pp. \bibinfo{pages}{230--239}.
  \bibinfo{note}{\url{https://doi.org/10.1145/3408308.3427617}}.
\bibitem[{Mocanu et~al.(2018)Mocanu, Mocanu, Nguyen, Liotta, Webber, Gibescu,
  and Slootweg}]{mocanu2018line}
\bibinfo{author}{E.~Mocanu}, \bibinfo{author}{D.~C. Mocanu},
  \bibinfo{author}{P.~H. Nguyen}, \bibinfo{author}{A.~Liotta},
  \bibinfo{author}{M.~E. Webber}, \bibinfo{author}{M.~Gibescu},
  \bibinfo{author}{J.~G. Slootweg},
\newblock \bibinfo{title}{On-line building energy optimization using deep
  reinforcement learning},
\newblock \bibinfo{journal}{IEEE transactions on smart grid}
  \bibinfo{volume}{10} (\bibinfo{year}{2018}) \bibinfo{pages}{3698--3708}.
  \bibinfo{note}{\url{https://doi.org/10.1109/TSG.2018.2834219}}.
\bibitem[{{Wei} et~al.(2017){Wei}, {Yanzhi Wang}, and {Zhu}}]{8060306}
\bibinfo{author}{T.~{Wei}}, \bibinfo{author}{{Yanzhi Wang}},
  \bibinfo{author}{Q.~{Zhu}},
\newblock \bibinfo{title}{Deep reinforcement learning for building hvac
  control},
\newblock in: \bibinfo{booktitle}{2017 54th ACM/EDAC/IEEE Design Automation
  Conference (DAC)}, pp. \bibinfo{pages}{1--6}.
  \bibinfo{note}{\url{https://doi.org/10.1145/3061639.3062224}}.
\bibitem[{Chen et~al.(2018)Chen, Norford, Samuelson, and Malkawi}]{CHEN2018195}
\bibinfo{author}{Y.~Chen}, \bibinfo{author}{L.~K. Norford},
  \bibinfo{author}{H.~W. Samuelson}, \bibinfo{author}{A.~Malkawi},
\newblock \bibinfo{title}{Optimal control of hvac and window systems for
  natural ventilation through reinforcement learning},
\newblock \bibinfo{journal}{Energy and Buildings} \bibinfo{volume}{169}
  (\bibinfo{year}{2018}) \bibinfo{pages}{195 -- 205}.
  \bibinfo{note}{\url{https://doi.org/10.1016/j.enbuild.2018.03.051}}.
\bibitem[{{Chen} et~al.(2017){Chen}, {Shi}, and {Zhang}}]{2017arXiv171102278C}
\bibinfo{author}{Y.~{Chen}}, \bibinfo{author}{Y.~{Shi}},
  \bibinfo{author}{B.~{Zhang}},
\newblock \bibinfo{title}{{Modeling and Optimization of Complex Building Energy
  Systems with Deep Neural Networks}},
\newblock \bibinfo{journal}{arXiv e-prints}  (\bibinfo{year}{2017})
  \bibinfo{pages}{arXiv:1711.02278}.
  \bibinfo{note}{\url{https://ui.adsabs.harvard.edu/abs/2017arXiv171102278C}}.
\bibitem[{Afram et~al.(2017)Afram, Janabi-Sharifi, Fung, and
  Raahemifar}]{AFRAM201796}
\bibinfo{author}{A.~Afram}, \bibinfo{author}{F.~Janabi-Sharifi},
  \bibinfo{author}{A.~S. Fung}, \bibinfo{author}{K.~Raahemifar},
\newblock \bibinfo{title}{Artificial neural network (ann) based model
  predictive control (mpc) and optimization of hvac systems: A state of the art
  review and case study of a residential hvac system},
\newblock \bibinfo{journal}{Energy and Buildings} \bibinfo{volume}{141}
  (\bibinfo{year}{2017}) \bibinfo{pages}{96 -- 113}. \bibinfo{note}{\url{
  https://doi.org/10.1016/j.enbuild.2017.02.012}}.
\bibitem[{Wang et~al.(2017)Wang, Velswamy, and Huang}]{pr5030046}
\bibinfo{author}{Y.~Wang}, \bibinfo{author}{K.~Velswamy},
  \bibinfo{author}{B.~Huang},
\newblock \bibinfo{title}{A long-short term memory recurrent neural network
  based reinforcement learning controller for office heating ventilation and
  air conditioning systems},
\newblock \bibinfo{journal}{Processes} \bibinfo{volume}{5}
  (\bibinfo{year}{2017}).
  \bibinfo{note}{\url{https://doi.org/10.3390/pr5030046}}.
\bibitem[{{Ruelens} et~al.(2017){Ruelens}, {Claessens}, {Vandael}, {De
  Schutter}, {Babuška}, and {Belmans}}]{7401112}
\bibinfo{author}{F.~{Ruelens}}, \bibinfo{author}{B.~J. {Claessens}},
  \bibinfo{author}{S.~{Vandael}}, \bibinfo{author}{B.~{De Schutter}},
  \bibinfo{author}{R.~{Babuška}}, \bibinfo{author}{R.~{Belmans}},
\newblock \bibinfo{title}{Residential demand response of thermostatically
  controlled loads using batch reinforcement learning},
\newblock \bibinfo{journal}{IEEE Transactions on Smart Grid}
  \bibinfo{volume}{8} (\bibinfo{year}{2017}) \bibinfo{pages}{2149--2159}.
  \bibinfo{note}{\url{https://doi.org/10.1109/TSG.2016.2517211}}.
\bibitem[{{Wenbo Shi} and {Wong}(2011)}]{6102330}
\bibinfo{author}{{Wenbo Shi}}, \bibinfo{author}{V.~W.~S. {Wong}},
\newblock \bibinfo{title}{Real-time vehicle-to-grid control algorithm under
  price uncertainty},
\newblock in: \bibinfo{booktitle}{2011 IEEE International Conference on Smart
  Grid Communications (SmartGridComm)}, pp. \bibinfo{pages}{261--266}.
  \bibinfo{note}{\url{https://doi.org/10.1109/SmartGridComm.2011.6102330}}.
\bibitem[{{Chiş} et~al.(2015){Chiş}, {Lundén}, and {Koivunen}}]{7178338}
\bibinfo{author}{A.~{Chiş}}, \bibinfo{author}{J.~{Lundén}},
  \bibinfo{author}{V.~{Koivunen}},
\newblock \bibinfo{title}{Optimization of plug-in electric vehicle charging
  with forecasted price},
\newblock in: \bibinfo{booktitle}{2015 IEEE International Conference on
  Acoustics, Speech and Signal Processing (ICASSP)}, pp.
  \bibinfo{pages}{2086--2089}.
  \bibinfo{note}{\url{https://doi.org/10.1109/ICASSP.2015.7178338}}.
\bibitem[{{Chiş} et~al.(2013){Chiş}, {Lundén}, and {Koivunen}}]{6695263}
\bibinfo{author}{A.~{Chiş}}, \bibinfo{author}{J.~{Lundén}},
  \bibinfo{author}{V.~{Koivunen}},
\newblock \bibinfo{title}{Scheduling of plug-in electric vehicle battery
  charging with price prediction},
\newblock in: \bibinfo{booktitle}{IEEE PES ISGT Europe 2013}, pp.
  \bibinfo{pages}{1--5}.
  \bibinfo{note}{\url{https://doi.org/10.1109/ISGTEurope.2013.6695263}}.
\bibitem[{{Sadeghianpourhamami} et~al.(2019){Sadeghianpourhamami}, {Deleu}, and
  {Develder}}]{8727484}
\bibinfo{author}{N.~{Sadeghianpourhamami}}, \bibinfo{author}{J.~{Deleu}},
  \bibinfo{author}{C.~{Develder}},
\newblock \bibinfo{title}{Definition and evaluation of model-free coordination
  of electrical vehicle charging with reinforcement learning},
\newblock \bibinfo{journal}{IEEE Transactions on Smart Grid}
  (\bibinfo{year}{2019}) \bibinfo{pages}{1--1}.
  \bibinfo{note}{\url{https://doi.org/10.1109/TSG.2019.2920320},
  ISSN={1949-3053}, month={},}.
\bibitem[{{Ko} et~al.(2018){Ko}, {Pack}, and {Leung}}]{8335743}
\bibinfo{author}{H.~{Ko}}, \bibinfo{author}{S.~{Pack}},
  \bibinfo{author}{V.~C.~M. {Leung}},
\newblock \bibinfo{title}{Mobility-aware vehicle-to-grid control algorithm in
  microgrids},
\newblock \bibinfo{journal}{IEEE Transactions on Intelligent Transportation
  Systems} \bibinfo{volume}{19} (\bibinfo{year}{2018})
  \bibinfo{pages}{2165--2174}.
  \bibinfo{note}{\url{https://doi.org/10.1109/TITS.2018.2816935},
  ISSN={1524-9050}, month={July},}.
\bibitem[{{Vandael} et~al.(2015){Vandael}, {Claessens}, {Ernst}, {Holvoet}, and
  {Deconinck}}]{7056534}
\bibinfo{author}{S.~{Vandael}}, \bibinfo{author}{B.~{Claessens}},
  \bibinfo{author}{D.~{Ernst}}, \bibinfo{author}{T.~{Holvoet}},
  \bibinfo{author}{G.~{Deconinck}},
\newblock \bibinfo{title}{Reinforcement learning of heuristic ev fleet charging
  in a day-ahead electricity market},
\newblock \bibinfo{journal}{IEEE Transactions on Smart Grid}
  \bibinfo{volume}{6} (\bibinfo{year}{2015}) \bibinfo{pages}{1795--1805}.
  \bibinfo{note}{\url{https://doi.org/10.1109/TSG.2015.2393059}}.
\bibitem[{Kim and Lim(2018)}]{en11082010}
\bibinfo{author}{S.~Kim}, \bibinfo{author}{H.~Lim},
\newblock \bibinfo{title}{Reinforcement learning based energy management
  algorithm for smart energy buildings},
\newblock \bibinfo{journal}{Energies} \bibinfo{volume}{11}
  (\bibinfo{year}{2018}).
  \bibinfo{note}{\url{https://www.mdpi.com/1996-1073/11/8/2010}}.
\bibitem[{{Nguyen} et~al.(2015){Nguyen}, {Nguyen}, and {Le}}]{6892986}
\bibinfo{author}{H.~T. {Nguyen}}, \bibinfo{author}{D.~T. {Nguyen}},
  \bibinfo{author}{L.~B. {Le}},
\newblock \bibinfo{title}{Energy management for households with solar assisted
  thermal load considering renewable energy and price uncertainty},
\newblock \bibinfo{journal}{IEEE Transactions on Smart Grid}
  \bibinfo{volume}{6} (\bibinfo{year}{2015}) \bibinfo{pages}{301--314}.
  \bibinfo{note}{\url{https://doi.org/10.1109/TSG.2014.2350831}}.
\bibitem[{{Kim} et~al.(2013){Kim}, {Ren}, {van der Schaar}, and
  {Lee}}]{6547831}
\bibinfo{author}{B.~{Kim}}, \bibinfo{author}{S.~{Ren}},
  \bibinfo{author}{M.~{van der Schaar}}, \bibinfo{author}{J.~{Lee}},
\newblock \bibinfo{title}{Bidirectional energy trading and residential load
  scheduling with electric vehicles in the smart grid},
\newblock \bibinfo{journal}{IEEE Journal on Selected Areas in Communications}
  \bibinfo{volume}{31} (\bibinfo{year}{2013}) \bibinfo{pages}{1219--1234}.
  \bibinfo{note}{\url{https://doi.org/10.1109/JSAC.2013.130706}}.
\bibitem[{{Nguyen} and {Le}(2014)}]{6596523}
\bibinfo{author}{D.~T. {Nguyen}}, \bibinfo{author}{L.~B. {Le}},
\newblock \bibinfo{title}{Joint optimization of electric vehicle and home
  energy scheduling considering user comfort preference},
\newblock \bibinfo{journal}{IEEE Transactions on Smart Grid}
  \bibinfo{volume}{5} (\bibinfo{year}{2014}) \bibinfo{pages}{188--199}.
  \bibinfo{note}{\url{10.1109/TSG.2013.2274521}}.
\bibitem[{noa(2021{\natexlab{a}})}]{noauthor_dfab_nodate}
\bibinfo{title}{{DFAB} {HOUSE} – {Digital} {Fabrication} and {Living},
  {Empa}, {Duebendorf}, {Switzerland}}, \bibinfo{year}{2021}{\natexlab{a}}.
  \bibinfo{note}{\url{https://www.empa.ch/web/nest/digital-fabrication},
  (Accessed: 20.05.2021.)}.
\bibitem[{noa(2021{\natexlab{b}})}]{noauthor_nest_nodate}
\bibinfo{title}{Nest – {Exploring} the {Future} of {Buildings}, {Swiss}
  {Federal} {Laboratories} for {Materials} {Science} and {Technology} - {EMPA},
  {Duebendorf}, {Switzerland}}, \bibinfo{year}{2021}{\natexlab{b}}.
  \bibinfo{note}{\url{https://www.empa.ch/web/nest}, (Accessed: 20.05.2021.)}.
\bibitem[{Keller(2021)}]{dfab_img}
\bibinfo{author}{R.~Keller}, \bibinfo{title}{Dfab house}, \bibinfo{year}{2021}.
  \bibinfo{note}{\url
  {https://www.swiss-architects.com/de/architecture-news/bau-der-woche/building-digitally-living-digitally?utm_source=newsletter&utm_medium=email&utm_campaign=2417},
  (Accessed 20.05.2021.)}.
\bibitem[{Lipton et~al.(2015)Lipton, Berkowitz, and Elkan}]{lipton2015critical}
\bibinfo{author}{Z.~C. Lipton}, \bibinfo{author}{J.~Berkowitz},
  \bibinfo{author}{C.~Elkan},
\newblock \bibinfo{title}{A critical review of recurrent neural networks for
  sequence learning},
\newblock \bibinfo{journal}{arXiv preprint arXiv:1506.00019}
  (\bibinfo{year}{2015}).
  \bibinfo{note}{\url{https://arxiv.org/abs/1506.00019}}.
\bibitem[{{Kingma} and {Ba}(2014)}]{adam}
\bibinfo{author}{D.~P. {Kingma}}, \bibinfo{author}{J.~{Ba}},
\newblock \bibinfo{title}{{Adam: A Method for Stochastic Optimization}},
\newblock \bibinfo{journal}{arXiv e-prints}  (\bibinfo{year}{2014})
  \bibinfo{pages}{arXiv:1412.6980}.
  \bibinfo{note}{\url{https://ui.adsabs.harvard.edu/abs/2014arXiv1412.6980K}}.
\bibitem[{Bergstra et~al.(2011)Bergstra, Bardenet, Bengio, and
  K\'{e}gl}]{hyper_tpe}
\bibinfo{author}{J.~Bergstra}, \bibinfo{author}{R.~Bardenet},
  \bibinfo{author}{Y.~Bengio}, \bibinfo{author}{B.~K\'{e}gl},
\newblock \bibinfo{title}{Algorithms for hyper-parameter optimization},
\newblock in: \bibinfo{booktitle}{Proceedings of the 24th International
  Conference on Neural Information Processing Systems}, NIPS’11,
  \bibinfo{publisher}{Curran Associates Inc.}, \bibinfo{address}{Red Hook, NY,
  USA}, \bibinfo{year}{2011}, p. \bibinfo{pages}{2546–2554}.
  \bibinfo{note}{\url{https://hal.inria.fr/hal-00642998/}}.
\bibitem[{Bergstra et~al.(2015)Bergstra, Komer, Eliasmith, Yamins, and
  Cox}]{bergstra2015hyperopt}
\bibinfo{author}{J.~Bergstra}, \bibinfo{author}{B.~Komer},
  \bibinfo{author}{C.~Eliasmith}, \bibinfo{author}{D.~Yamins},
  \bibinfo{author}{D.~D. Cox},
\newblock \bibinfo{title}{Hyperopt: a python library for model selection and
  hyperparameter optimization},
\newblock \bibinfo{journal}{Computational Science \& Discovery}
  \bibinfo{volume}{8} (\bibinfo{year}{2015}) \bibinfo{pages}{014008}.
  \bibinfo{note}{\url{https://doi.org/10.1088/1749-4699/8/1/014008}}.
\bibitem[{Dow(2020)}]{dow_tesla_2020}
\bibinfo{author}{J.~Dow}, \bibinfo{title}{Tesla releases ‘long range plus’
  model s/x with 390/351 mile range, new wheels}, \bibinfo{year}{2020}.
  \bibinfo{note}{\url
  {https://electrek.co/2020/02/14/tesla-releases-long-range-plus-model-s-x-with-390-351-mile-range-new-wheels/},
  (Accessed: 20.05.2021.)}.
\bibitem[{Brodie(2018)}]{brodie_autoexpress_2018}
\bibinfo{author}{J.~Brodie}, \bibinfo{title}{{AutoExpress}: {BMW} i3 updated
  with more range and new trim options}, \bibinfo{year}{2018}.
  \bibinfo{note}{\url
  {https://www.autoexpress.co.uk/bmw/i3/104793/bmw-i3-updated-with-more-range-and-new-trim-options},
  (Accessed: 20.05.2021.)}.
\bibitem[{Taylor(2018)}]{taylor_better_2018}
\bibinfo{author}{M.~Taylor}, \bibinfo{title}{Better late than never as {EQC}
  leads {Mercedes-Benz} {EV} assault}, \bibinfo{year}{2018}.
  \bibinfo{note}{\url
  {https://www.forbes.com/sites/michaeltaylor/2018/09/04/better-late-than-never-as-eqc-leads-mercedes-benz-ev-assault/?sh=25d6245c5f5a},
  (Accessed: 20.05.2021.)}.
\bibitem[{{Qiu} et~al.(2019){Qiu}, {Hu}, {Chen}, and {Zeng}}]{ddpg_app_1}
\bibinfo{author}{C.~{Qiu}}, \bibinfo{author}{Y.~{Hu}},
  \bibinfo{author}{Y.~{Chen}}, \bibinfo{author}{B.~{Zeng}},
\newblock \bibinfo{title}{Deep deterministic policy gradient {(DDPG)}-based
  energy harvesting wireless communications},
\newblock \bibinfo{journal}{IEEE Internet of Things Journal}
  \bibinfo{volume}{6} (\bibinfo{year}{2019}) \bibinfo{pages}{8577--8588}.
  \bibinfo{note}{\url{https://doi.org/10.1109/JIOT.2019.2921159}}.
\bibitem[{{Vecerik} et~al.(2017){Vecerik}, {Hester}, {Scholz}, {Wang},
  {Pietquin}, {Piot}, {Heess}, {Roth{\"o}rl}, {Lampe}, and
  {Riedmiller}}]{2017arXiv170708817V}
\bibinfo{author}{M.~{Vecerik}}, \bibinfo{author}{T.~{Hester}},
  \bibinfo{author}{J.~{Scholz}}, \bibinfo{author}{F.~{Wang}},
  \bibinfo{author}{O.~{Pietquin}}, \bibinfo{author}{B.~{Piot}},
  \bibinfo{author}{N.~{Heess}}, \bibinfo{author}{T.~{Roth{\"o}rl}},
  \bibinfo{author}{T.~{Lampe}}, \bibinfo{author}{M.~{Riedmiller}},
\newblock \bibinfo{title}{{Leveraging Demonstrations for Deep Reinforcement
  Learning on Robotics Problems with Sparse Rewards}},
\newblock \bibinfo{journal}{arXiv e-prints}  (\bibinfo{year}{2017})
  \bibinfo{pages}{arXiv:1707.08817}.
  \bibinfo{note}{\url{https://ui.adsabs.harvard.edu/abs/2017arXiv170708817V}}.
\bibitem[{{Xiong} et~al.(2016){Xiong}, {Wang}, {Zhang}, and
  {Li}}]{2016arXiv161200147X}
\bibinfo{author}{X.~{Xiong}}, \bibinfo{author}{J.~{Wang}},
  \bibinfo{author}{F.~{Zhang}}, \bibinfo{author}{K.~{Li}},
\newblock \bibinfo{title}{{Combining Deep Reinforcement Learning and Safety
  Based Control for Autonomous Driving}},
\newblock \bibinfo{journal}{arXiv e-prints}  (\bibinfo{year}{2016})
  \bibinfo{pages}{arXiv:1612.00147}.
  \bibinfo{note}{\url{https://ui.adsabs.harvard.edu/abs/2016arXiv161200147X}}.
\bibitem[{{Yang} et~al.(2019){Yang}, {Zhu}, {Zhang}, {Qiao}, and
  {Liu}}]{ddpg_app_2}
\bibinfo{author}{Q.~{Yang}}, \bibinfo{author}{Y.~{Zhu}},
  \bibinfo{author}{J.~{Zhang}}, \bibinfo{author}{S.~{Qiao}},
  \bibinfo{author}{J.~{Liu}},
\newblock \bibinfo{title}{{UAV} air combat autonomous maneuver decision based
  on {DDPG} algorithm},
\newblock in: \bibinfo{booktitle}{2019 IEEE 15th International Conference on
  Control and Automation (ICCA)}, pp. \bibinfo{pages}{37--42}.
  \bibinfo{note}{\url{https://doi.org/10.1109/ICCA.2019.8899703}}.
\bibitem[{Chollet et~al.(2015)}]{chollet2015keras}
\bibinfo{author}{F.~Chollet}, et~al., \bibinfo{title}{Keras},
  \bibinfo{howpublished}{\url{https://keras.io}}, \bibinfo{year}{2015}.
\bibitem[{Plappert(2016)}]{plappert2016kerasrl}
\bibinfo{author}{M.~Plappert}, \bibinfo{title}{keras-rl},
  \bibinfo{howpublished}{\url{https://github.com/keras-rl/keras-rl}},
  \bibinfo{year}{2016}.
\bibitem[{Finch(2004)}]{Finch04ouprocess}
\bibinfo{author}{S.~Finch}, \bibinfo{title}{Ornstein-uhlenbeck process},
  \bibinfo{year}{2004}.
  \bibinfo{note}{\url{http://citeseerx.ist.psu.edu/viewdoc/summary?doi=10.1.1.710.4200}}.
\bibitem[{Pedregosa et~al.(2011)Pedregosa, Varoquaux, Gramfort, Michel,
  Thirion, Grisel, Blondel, Prettenhofer, Weiss, Dubourg, Vanderplas, Passos,
  Cournapeau, Brucher, Perrot, and Duchesnay}]{scikit-learn}
\bibinfo{author}{F.~Pedregosa}, \bibinfo{author}{G.~Varoquaux},
  \bibinfo{author}{A.~Gramfort}, \bibinfo{author}{V.~Michel},
  \bibinfo{author}{B.~Thirion}, \bibinfo{author}{O.~Grisel},
  \bibinfo{author}{M.~Blondel}, \bibinfo{author}{P.~Prettenhofer},
  \bibinfo{author}{R.~Weiss}, \bibinfo{author}{V.~Dubourg},
  \bibinfo{author}{J.~Vanderplas}, \bibinfo{author}{A.~Passos},
  \bibinfo{author}{D.~Cournapeau}, \bibinfo{author}{M.~Brucher},
  \bibinfo{author}{M.~Perrot}, \bibinfo{author}{E.~Duchesnay},
\newblock \bibinfo{title}{Scikit-learn: Machine learning in {P}ython},
\newblock \bibinfo{journal}{Journal of Machine Learning Research}
  \bibinfo{volume}{12} (\bibinfo{year}{2011}) \bibinfo{pages}{2825--2830}.
  \bibinfo{note}{\url{https://www.jmlr.org/papers/volume12/pedregosa11a/pedregosa11a.pdf?source=post_page---------------------------}}.
\bibitem[{De~Coninck and Helsen(2016)}]{de2016practical}
\bibinfo{author}{R.~De~Coninck}, \bibinfo{author}{L.~Helsen},
\newblock \bibinfo{title}{Practical implementation and evaluation of model
  predictive control for an office building in brussels},
\newblock \bibinfo{journal}{Energy and Buildings} \bibinfo{volume}{111}
  (\bibinfo{year}{2016}) \bibinfo{pages}{290--298}.
  \bibinfo{note}{\url{https://doi.org/10.1016/j.enbuild.2015.11.014}}.
\bibitem[{Abadi et~al.(2015)Abadi, Agarwal, Barham, Brevdo, Chen, Citro,
  Corrado, Davis, Dean, Devin, Ghemawat, Goodfellow, Harp, Irving, Isard, Jia,
  Jozefowicz, Kaiser, Kudlur, Levenberg, Man\'{e}, Monga, Moore, Murray, Olah,
  Schuster, Shlens, Steiner, Sutskever, Talwar, Tucker, Vanhoucke, Vasudevan,
  Vi\'{e}gas, Vinyals, Warden, Wattenberg, Wicke, Yu, and
  Zheng}]{tensorflow2015}
\bibinfo{author}{M.~Abadi}, \bibinfo{author}{A.~Agarwal},
  \bibinfo{author}{P.~Barham}, \bibinfo{author}{E.~Brevdo},
  \bibinfo{author}{Z.~Chen}, \bibinfo{author}{C.~Citro}, \bibinfo{author}{G.~S.
  Corrado}, \bibinfo{author}{A.~Davis}, \bibinfo{author}{J.~Dean},
  \bibinfo{author}{M.~Devin}, \bibinfo{author}{S.~Ghemawat},
  \bibinfo{author}{I.~Goodfellow}, \bibinfo{author}{A.~Harp},
  \bibinfo{author}{G.~Irving}, \bibinfo{author}{M.~Isard},
  \bibinfo{author}{Y.~Jia}, \bibinfo{author}{R.~Jozefowicz},
  \bibinfo{author}{L.~Kaiser}, \bibinfo{author}{M.~Kudlur},
  \bibinfo{author}{J.~Levenberg}, \bibinfo{author}{D.~Man\'{e}},
  \bibinfo{author}{R.~Monga}, \bibinfo{author}{S.~Moore},
  \bibinfo{author}{D.~Murray}, \bibinfo{author}{C.~Olah},
  \bibinfo{author}{M.~Schuster}, \bibinfo{author}{J.~Shlens},
  \bibinfo{author}{B.~Steiner}, \bibinfo{author}{I.~Sutskever},
  \bibinfo{author}{K.~Talwar}, \bibinfo{author}{P.~Tucker},
  \bibinfo{author}{V.~Vanhoucke}, \bibinfo{author}{V.~Vasudevan},
  \bibinfo{author}{F.~Vi\'{e}gas}, \bibinfo{author}{O.~Vinyals},
  \bibinfo{author}{P.~Warden}, \bibinfo{author}{M.~Wattenberg},
  \bibinfo{author}{M.~Wicke}, \bibinfo{author}{Y.~Yu},
  \bibinfo{author}{X.~Zheng}, \bibinfo{title}{{TensorFlow}: Large-scale machine
  learning on heterogeneous systems}, \bibinfo{year}{2015}.
  \bibinfo{note}{\url{https://arxiv.org/abs/1603.04467}}.
\bibitem[{Oliphant(2006)}]{numpy}
\bibinfo{author}{T.~E. Oliphant}, \bibinfo{title}{A guide to NumPy},
  volume~\bibinfo{volume}{1}, \bibinfo{publisher}{Trelgol Publishing USA},
  \bibinfo{year}{2006}.
  \bibinfo{note}{\url{https://ecs.wgtn.ac.nz/foswiki/pub/Support/ManualPagesAndDocumentation/numpybook.pdf}}.
\bibitem[{McKinney(2011)}]{mckinney2011pandas}
\bibinfo{author}{W.~McKinney},
\newblock \bibinfo{title}{pandas: a foundational python library for data
  analysis and statistics},
\newblock \bibinfo{journal}{Python for High Performance and Scientific
  Computing} \bibinfo{volume}{14} (\bibinfo{year}{2011}).
  \bibinfo{note}{\url{https://www.dlr.de/sc/portaldata/15/resources/dokumente/pyhpc2011/submissions/pyhpc2011_submission_9.pdf}}.
\bibitem[{{Hunter}(2007)}]{mpl}
\bibinfo{author}{J.~D. {Hunter}},
\newblock \bibinfo{title}{Matplotlib: A 2d graphics environment},
\newblock \bibinfo{journal}{Computing in Science Engineering}
  \bibinfo{volume}{9} (\bibinfo{year}{2007}) \bibinfo{pages}{90--95}.
  \bibinfo{note}{\url{https://doi.org/10.1109/MCSE.2007.55}}.
\bibitem[{Brockman et~al.(2016)Brockman, Cheung, Pettersson, Schneider,
  Schulman, Tang, and Zaremba}]{1606.01540}
\bibinfo{author}{G.~Brockman}, \bibinfo{author}{V.~Cheung},
  \bibinfo{author}{L.~Pettersson}, \bibinfo{author}{J.~Schneider},
  \bibinfo{author}{J.~Schulman}, \bibinfo{author}{J.~Tang},
  \bibinfo{author}{W.~Zaremba}, \bibinfo{title}{Openai gym},
  \bibinfo{year}{2016}. \bibinfo{note}{\url{https://arxiv.org/abs/1606.01540}}.
\bibitem[{{Virtanen} et~al.(2020){Virtanen}, {Gommers}, {Oliphant},
  {Haberland}, {Reddy}, {Cournapeau}, {Burovski}, {Peterson}, {Weckesser},
  {Bright}, {van der Walt}, {Brett}, {Wilson}, {Jarrod Millman}, {Mayorov},
  {Nelson}, {Jones}, {Kern}, {Larson}, {Carey}, {Polat}, {Feng}, {Moore}, {Vand
  erPlas}, {Laxalde}, {Perktold}, {Cimrman}, {Henriksen}, {Quintero}, {Harris},
  {Archibald}, {Ribeiro}, {Pedregosa}, {van Mulbregt}, and
  {Contributors}}]{2020SciPy}
\bibinfo{author}{P.~{Virtanen}}, \bibinfo{author}{R.~{Gommers}},
  \bibinfo{author}{T.~E. {Oliphant}}, \bibinfo{author}{M.~{Haberland}},
  \bibinfo{author}{T.~{Reddy}}, \bibinfo{author}{D.~{Cournapeau}},
  \bibinfo{author}{E.~{Burovski}}, \bibinfo{author}{P.~{Peterson}},
  \bibinfo{author}{W.~{Weckesser}}, \bibinfo{author}{J.~{Bright}},
  \bibinfo{author}{S.~J. {van der Walt}}, \bibinfo{author}{M.~{Brett}},
  \bibinfo{author}{J.~{Wilson}}, \bibinfo{author}{K.~{Jarrod Millman}},
  \bibinfo{author}{N.~{Mayorov}}, \bibinfo{author}{A.~R.~J. {Nelson}},
  \bibinfo{author}{E.~{Jones}}, \bibinfo{author}{R.~{Kern}},
  \bibinfo{author}{E.~{Larson}}, \bibinfo{author}{C.~{Carey}},
  \bibinfo{author}{{\.I}.~{Polat}}, \bibinfo{author}{Y.~{Feng}},
  \bibinfo{author}{E.~W. {Moore}}, \bibinfo{author}{J.~{Vand erPlas}},
  \bibinfo{author}{D.~{Laxalde}}, \bibinfo{author}{J.~{Perktold}},
  \bibinfo{author}{R.~{Cimrman}}, \bibinfo{author}{I.~{Henriksen}},
  \bibinfo{author}{E.~A. {Quintero}}, \bibinfo{author}{C.~R. {Harris}},
  \bibinfo{author}{A.~M. {Archibald}}, \bibinfo{author}{A.~H. {Ribeiro}},
  \bibinfo{author}{F.~{Pedregosa}}, \bibinfo{author}{P.~{van Mulbregt}},
  \bibinfo{author}{S.~.~. {Contributors}},
\newblock \bibinfo{title}{{SciPy 1.0: Fundamental Algorithms for Scientific
  Computing in Python}},
\newblock \bibinfo{journal}{Nature Methods}  (\bibinfo{year}{2020}).
  \bibinfo{note}{\url {https://doi.org/10.1038/s41592-019-0686-2}}.
\bibitem[{Seabold and Perktold(2010)}]{seabold2010statsmodels}
\bibinfo{author}{S.~Seabold}, \bibinfo{author}{J.~Perktold},
\newblock \bibinfo{title}{Statsmodels: Econometric and statistical modeling
  with python},
\newblock in: \bibinfo{booktitle}{9th Python in Science Conference}.
  \bibinfo{note}{\url{https://pdfs.semanticscholar.org/3a27/6417e5350e29cb6bf04ea5a4785601d5a215.pdf}}.
\bibitem[{Rossum(1995)}]{python}
\bibinfo{author}{G.~Rossum}, \bibinfo{title}{Python Reference Manual},
  \bibinfo{type}{Technical Report}, \bibinfo{address}{NLD},
  \bibinfo{year}{1995}. \bibinfo{note}{\url{
  https://dl.acm.org/doi/abs/10.5555/869369}}.

\end{thebibliography}
